\documentclass[lettersize,journal]{IEEEtran}
\usepackage{amsmath,amsfonts}
\usepackage{algorithmic}
\usepackage{algorithm}
\usepackage{array}
\usepackage[caption=false,font=normalsize,labelfont=sf,textfont=sf]{subfig}
\usepackage{textcomp}
\usepackage{stfloats}
\usepackage{url}
\usepackage{verbatim}
\usepackage{graphicx}
\usepackage{cite}
\usepackage{orcidlink}
\usepackage{makecell}
\usepackage{tabularx}
\usepackage{multirow}
\usepackage{xcolor}
\usepackage{booktabs}
\usepackage{subcaption}
\usepackage{array} 
\usepackage{tabularx} 
\usepackage{makecell}
\usepackage{booktabs}

\hypersetup{hidelinks, colorlinks, linkcolor=blue, citecolor=blue, urlcolor=blue}
\newcommand{\mypara}[1]{{\vspace{0.1mm}\noindent\textbf{#1}}}

\newcommand\revise[1]{\textcolor{black}{}\textcolor{black}{#1}}

\begin{document}

\title{Multi-Modal Anomaly Detection: A Survey}

\author{Xudong Mou$^{\orcidlink{0009-0005-1445-3742}}$, 
Zexin Wu$^{\orcidlink{0009-0008-3772-0542}}$, 
Chuan Luo$^{\orcidlink{0000-0001-5028-1064}}$, 
Shiru Chen$^{\orcidlink{0009-0009-6870-3611}}$, 
Xudong Liu$^{\orcidlink{0000-0001-8566-660X}}$,
Chunming Hu$^{\orcidlink{0000-0003-3473-9703}}$,
Renyu Yang$^{\orcidlink{0000-0001-6334-4925}}$

\thanks{Manuscript received December 29, 2025; revised May 25, 2026; accepted 13 July 2026. This work is supported in part by National Key
R\&D Program of China (Grant No. 2024YFB4505901), Beijing Natural Science Foundation (No. L241050), National Natural Science Foundation of China (Grant No. 62402024), and the Fundamental Research Funds for the Central Universities. (Corresponding author: Renyu Yang)}

\thanks{
Xudong Mou and Xudong Liu are with the School of Computer Science and Engineering, Beihang University, Beijing 100191, China. 
(Email: mxd@buaa.edu.cn, liuxd@buaa.edu.cn).
}

\thanks{
Zexin Wu, Chuan Luo, Chunming Hu, and Renyu Yang are with the School of Software, Beihang University, Beijing 100191, China (Email: zexinwu@buaa.edu.cn, chuanluo@buaa.edu.cn, hucm@buaa.edu.cn, renyuyang@buaa.edu.cn).
}

\thanks{
Shiru Chen is with the Shandong Inspur Intelligent Production Technology Co., Ltd, Jinan, 250101, China (Email: chenshiru@inspur.com).
}

}

\markboth{IEEE Transactions on Big Data, ~Vol.~x, No.~x, August~2026}%
{Xudong Mou \MakeLowercase{\textit{et al.}}: Multi-Modal Anomaly Detection: A Survey}

\IEEEpubid{0000--0000/00\$00.00~\copyright~2026 IEEE}

\maketitle

\begin{abstract}
Multi-Modal Anomaly Detection (MMAD) detects rare abnormal events from heterogeneous data sources and is increasingly used in safety- and reliability-critical applications such as industrial inspection and cybersecurity. Yet the literature is fragmented across domains and modality combinations, and existing surveys usually group methods by architecture rather than by how abnormality is defined and separated in multi-modal settings.
We survey MMAD from an assumption-driven perspective. We formalize the problem, identify five intrinsic characteristics underlying its core challenges, and organize prior work into two complementary paradigms. The first, \emph{normality-assumption} methods, models regularity via representation learning, cross-modal alignment, and knowledge enhancement. The second, \emph{anomaly-assumption} methods, sharpens decision boundaries through coarse-grained, structural, and semantic anomaly injection. We also investigate how foundation models are reshaping MMAD through scalable pretraining, flexible cross-modal transfer, and emerging reasoning capabilities.
Finally, we compile representative benchmarks and evaluation protocols across domains and highlight open problems and future directions for robust, adaptive, and interpretable MMAD systems.

\end{abstract}


\section{Introduction}
Anomaly Detection (AD) identifies observations that deviate from normal behavior~\cite{pang2021deep,yang2024generalized}.
Because many AD applications are safety- and reliability-critical—such as industrial inspection, healthcare, surveillance, finance, and cybersecurity—the field has remained active for decades~\cite{wang2025acbot,de2025explainable}.
Traditional methods typically use unimodal data and model normality in one observation space, but often fail when anomalies are subtle, partially observed, or modality-dependent.

With the rise of heterogeneous data sources, Multi-Modal Anomaly Detection (MMAD) has become a natural extension of classical AD.
By combining evidence from modalities such as images, time series, text, audio, and sensor signals, MMAD can detect abnormal events that appear normal in any single modality~\cite{lin2025survey}.
Multi-modal integration also improves robustness to noise, missing data, and partial failures, yielding a more accurate view of complex systems~\cite{costanzino2024multimodal}.
However, multimodality also reshapes the problem by introducing heterogeneity and alignment issues, and by exacerbating challenges such as granularity mismatch, semantic ambiguity, and evolving data distributions. \IEEEpubidadjcol

Several surveys have examined MMAD and related areas. Existing reviews cover industrial visual anomaly detection with RGB, 3D, and multi-modal inputs~\cite{lin2025survey}, multi-modal datasets for audio--visual event understanding~\cite{kumari2024multimedia}, and advances in multi-modal time-series analysis~\cite{jiang2025multi}.
Other domain-focused surveys study multi-modal event detection and representation learning from collaboration and fusion perspectives~\cite{xiao2022survey,wang2021survey}.
However, these works either target specific domains (e.g., RGB/3D industrial images, audio--visual benchmarks, time-series signals) or discuss general learning tools without offering a holistic view of MMAD.

Interest in applying Foundation Models (FMs) to MMAD is rapidly growing.
Pretrained on large-scale, diverse corpora, FMs provide strong representation learning and cross-modal alignment that are hard to achieve with task-specific architectures~\cite{ren2025foundation,ben5251498foundation}.
 This has enabled promising zero-shot and few-shot anomaly detection, especially when labeled anomalies are scarce~\cite{zhou2024anomalyclip,jeong2023winclip}.
 FM-based methods also support flexible fusion and transfer, reducing handcrafted pipelines and easing adaptation to new modalities or anomaly types~\cite{zhao2025unimmad,zhang2025towards,jiang2025anomagic}.
Nevertheless, existing FM-centric surveys largely describe generic AD pipelines (e.g., FM-as-encoder or FM-as-interpreter) without systematically linking multi-modal pretraining, cross-modal alignment, and FM-driven inference to MMAD’s specific challenges. Their taxonomies are thus weakly tied to the essence of anomaly detection, tending to emphasize architectural novelty over clear explanations of detection mechanisms.

\begin{figure*}[ht]
  \centering
  \includegraphics[width=\linewidth]{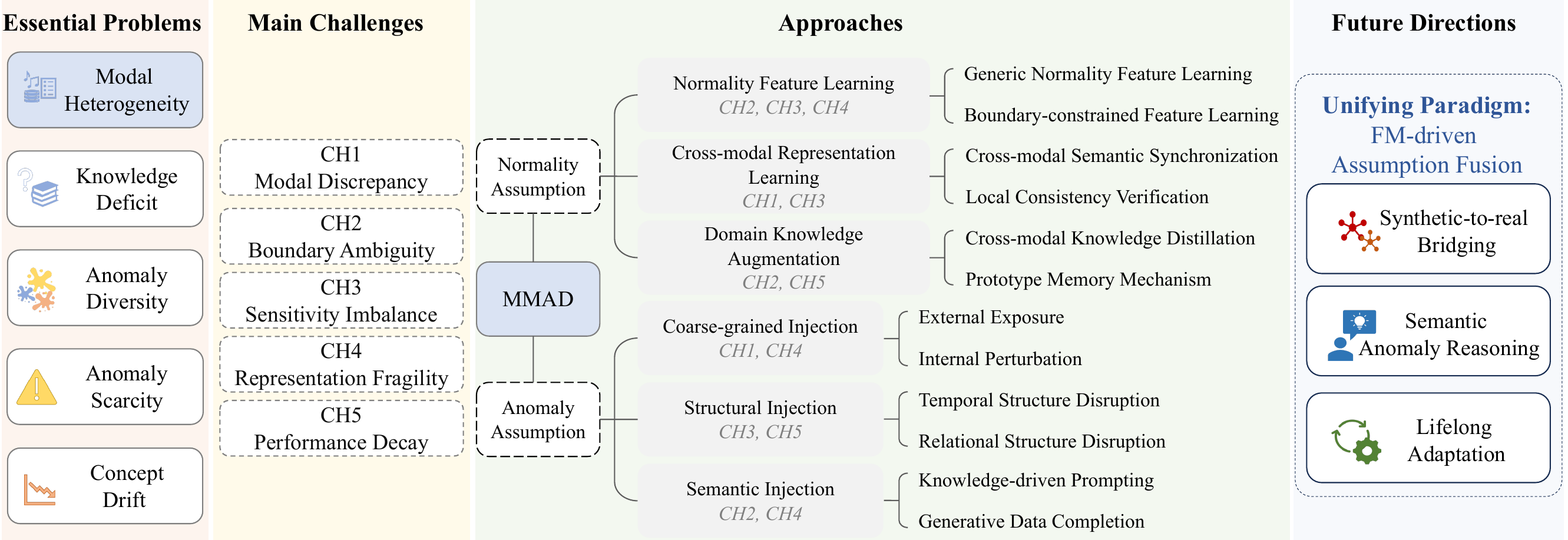}
  \caption{Conceptual roadmap of multi-modal anomaly detection.
The figure illustrates the progression from the intrinsic nature of anomaly detection to multi-modality-induced challenges, the assumption-driven modeling paradigms, and the emerging future opportunities in MMAD.}
  \label{architecture}
\end{figure*}

In this survey, we take an assumption-driven view of MMAD.
Instead of grouping methods by modality or model family, we organize them by their assumptions about \emph{normality} and \emph{abnormality}.
We review MMAD approaches through two lenses: (i) normality-based methods that model regular patterns and define anomalies implicitly, and (ii) abnormality-based methods that explicitly construct contrasts to mark what lies beyond normality. Within each lens, we trace the development from classical unimodal techniques to modern multi-modal methods, showing how their assumptions interact with MMAD-specific challenges.
Grounding the taxonomy in anomaly detection itself yields a unified view of how methods define, expose, and separate abnormal behavior in multi-modal settings.
This framework clarifies the strengths and limitations of existing approaches—including FM-based methods—and highlights open gaps and future research directions. Fig.~\ref{architecture} summarizes the roadmap of this survey.

The paper makes the following main contributions.

\begin{itemize}
\item To the best of our knowledge, we present the first systematic survey of MMAD from an assumption-driven perspective, providing a unified and interpretable view of the research landscape.
\item We analyze the intrinsic nature of MMAD and distill a set of fundamental challenges that explain why many problems remain only partially resolved.
\item We review a broad body of studies from leading conferences and journals through the complementary lenses of normality and abnormality assumptions, summarizing their objectives, capabilities, and limitations in addressing MMAD challenges.
\item We identify emerging trends and formulate future research directions for MMAD through the lens of foundation-model-driven assumption fusion, highlighting reliable anomaly hypothesis generation, semantic cross-modal reasoning, and lifelong adaptation.
\end{itemize}

The remainder of this paper is organized as follows. Section~\ref{sec:problem} defines MMAD and analyzes its core challenges. Sections~\ref{sec:normality} and~\ref{sec:anomaly} review assumption-driven methods from normality- and abnormality-centric perspectives, highlighting the impact of foundation models. Section~\ref{sec:datasets} summarizes benchmarks and applications, Section~\ref{sec:future} discusses open problems, and Section~\ref{sec:conclusion} concludes.

\section{Preliminaries}
\label{sec:problem}
\subsection{Definition of Multi-modal Anomaly Detection}

Multi-modal anomaly detection seeks to integrate diverse and heterogeneous information to identify rare anomalous samples that deviate substantially from the normal distribution~\cite{pang2021deep}.
By exploiting cross-modal complementarity, it can reveal complex anomalous patterns that may be obscured in unimodal views due to limited data.

\revise{
Let \(\mathcal{X} = \{\mathbf{x}^{(1)}, \dots, \mathbf{x}^{(K)}\}\) be a multi-modal sample with \(K\) heterogeneous modalities, where \(\mathbf{x}^{(k)} \in \mathbb{R}^{d_k}\) is the observation from modality \(k\) (e.g., text, images, time series, audio). To model missing modalities, let \(\mathcal{M} \subseteq \{1, \dots, K\}\) be the set of observed modalities, and define \(\mathcal{X}_{\mathcal{M}} := \{\mathbf{x}^{(k)}\}_{k \in \mathcal{M}}\).  
MMAD aims to learn an anomaly score \(\mathcal{S}\) via a scoring function \(f(\cdot)\) that measures how far a sample deviates from the joint normal distribution \(\mathcal{P}_{\text{normal}}\).
\begin{equation}
    \mathcal{S} = f(\mathcal{X}, \mathcal{M}) := f(\mathcal{X}_{\mathcal{M}}) \in \mathbb{R}.
\end{equation}
The decision rule for a test sample is:
\begin{equation}
    \text{Label}(\mathcal{X}, \mathcal{M}) =
    \begin{cases}
    1\ (\text{Anomaly}), & \text{if } \mathcal{S} > \tau, \\
    0\ (\text{Normal}), & \text{if } \mathcal{S} \le \tau,
    \end{cases}
\end{equation}
where \(\tau\) is a predefined threshold typically determined on a validation set.
}

\revise{
Compared with unimodal anomaly detection, MMAD must capture both within-modality regularities (in each $\mathbf{x}^{(k)}$) and cross-modal dependencies (across $\mathcal{M}$), introducing challenges such as cross-modal inconsistency, modality imbalance, and missing modalities.}

This formulation highlights three issues that deserve particular attention. 
(i) Since anomalies are defined relative to normality, multi-modal normality should be explicitly specified and aligned with the target task, capturing both within-modality regularities and cross-modal structural consistency; otherwise, detection results may be unreliable. 
(ii) Anomaly detection typically operates under extreme class imbalance, and in multi-modal scenarios, anomalies may arise from rare cross-modal interactions rather than from a single modality alone. 
(iii) Following \cite{yang2024generalized}, it is crucial to distinguish anomaly detection (AD), novelty detection (ND), outlier detection (OD), and out-of-distribution (OOD) detection, which differ in problem settings and assumptions and become more nuanced in multi-modal contexts due to interacting information sources. For example, AD detects deviations from established normal patterns, ND treats novel samples as potential new categories, OOD emphasizes robustness in multi-class settings, and OD identifies outliers within a given dataset.

These considerations underscore the inherent complexity of MMAD and motivate the detailed discussion of its key challenges in the next section.

\subsection{Essential Problems in MMAD}

The complexity of MMAD can be deconstructed into the inherent challenges of anomaly detection itself and the complications introduced by the integration of multiple data sources. We summarize these essential problems along two dimensions: the intrinsic nature of anomalies and the exacerbation introduced by multimodality.

\subsubsection{Intrinsic characteristics of anomaly detection} 
At its core, anomaly detection addresses the pursuit of the ``unknown" and the ``rare". 
The primary bottleneck is \textit{anomaly scarcity}: real-world anomalies are extremely infrequent, making representative datasets of failure modes nearly impossible to collect. 
This data poverty leads to a chronic \textit{knowledge deficit}, as the mechanisms, statistical distributions, and boundary conditions of anomalies remain poorly understood. 
\textit{Anomaly diversity} also appears in unimodal settings, making modeling inherently challenging, for example, point- and pattern-wise anomalies in time series and pixel- and patch-level anomalies in images.
\textit{Concept drift} further challenges any deployment, as the notion of normality continuously evolves in non-stationary environments.

\subsubsection{Exacerbation introduced by multimodality} 
Introducing multiple modalities brings \textit{modal heterogeneity}, where data from disparate sensors exhibit inconsistent distributions, asynchronous sampling, and contradictory signals. Importantly, multimodality amplifies the inherent challenges of AD. Knowledge deficits are worsened when scarce abnormal signals must be reconciled across unaligned feature spaces. Anomaly diversity extends to the cross-modal level, besides anomalies that cross modalities, including spatial, relational, temporal, or semantic, an event may appear normal within a single stream but manifest as anomalous through inter-modal conflicts. Concept drift also becomes multi-dimensional and harder to calibrate, as different modalities may evolve at varying rates or even in opposing directions.

In summary, the landscape of MMAD is defined by five key characteristics: modal heterogeneity, knowledge deficit, anomaly diversity, anomaly scarcity, and concept drift.

\subsection{Main Challenges in MMAD}
Despite recent progress in scaling and high-dimensional modeling, MMAD still faces several unresolved challenges.

\subsubsection{CH1: Modal discrepancy from heterogeneity}
Modalities such as text, images, time series, and audio exhibit substantial distributional heterogeneity in feature space~\cite{wang2023multimodal}.
Anomalies may manifest differently across modalities, and effective detection can require integrating multiple streams.
For example, during robotic grasping, a micro-slip induces distinct friction changes in force/torque time-series signals but remains visually imperceptible. Without accurate semantic alignment, the model may exhibit visual bias, miss the slip, and even cause collisions during assembly.
Achieving semantically aligned representations is therefore critical.
The goal is for modalities to complement one another, rather than acting as ``squabbling experts'' that produce conflicting predictions.

\subsubsection{CH2: Boundary ambiguity from knowledge deficit}
The boundary between normality and anomaly is often blurred by limited expert guidance and weak or noisy annotations.
A statistical outlier may be a rare yet legitimate operating state in specialized domains, complicating boundary optimization~\cite{peng2025adaptive}. Conversely, data collected during the wear-out stage of the bathtub curve may naturally include failure patterns, contaminating the ``normal'' baseline.
Although anomalies are frequently treated as unknowns, basic priors such as the $3\sigma$ rule provide indispensable constraints. Forcing a model to learn entirely from data without such domain priors is inefficient and often increases false-alarm rates.
Leveraging limited domain knowledge to form robust and precise decision boundaries thus remains a key challenge.

\subsubsection{CH3: Sensitivity imbalance from anomaly diversity}
Anomalies vary widely in scale, ranging from point-level outliers to long-horizon patterns and large, patchy structural deviations. This requires sensitivity across multiple temporal and spatial granularities.
For example, in intrusion detection, brute-force attacks can appear as high-frequency point anomalies, whereas low-and-slow exfiltration emerges over weeks as a subtle collective pattern.
Emphasizing only global features can miss local irregularities, while over-indexing on local noise risks overfitting. The goal is to develop multi-scale mechanisms that capture the anomaly spectrum, from micro-structural glitches to macro-level semantic violations.

\subsubsection{CH4: Representation fragility from anomaly scarcity}
Real-world anomalies are naturally scarce, creating severe class imbalance and overwhelming models with normal samples. This scarcity forces detectors to rely heavily on the normality assumption, often degrading recall for rare or unseen patterns.
For instance, in critical infrastructure monitoring, catastrophic failures may never occur in the observed history, yielding ``black swan'' events that models cannot extrapolate to.
The central challenge is to complete the flip side of the distribution to sharpen the decision boundary and reliably identify rare threats.

\subsubsection{CH5: Performance decay from concept drift}
In non-stationary environments, the notion of normality evolves, gradually eroding model efficacy. Drift arises when environmental changes or system aging alter data statistics, making previously learned normal manifolds obsolete.
For example, in industrial IoT, sensor readings for healthy machines can drift as components wear down or seasonal temperature shifts occur. Without adaptation mechanisms, models may flag these legitimate shifts as anomalies.
Developing lifelong learning capabilities that separate true anomalies from a new normality is therefore essential for sustained deployment.

\section{Methods based on Normality Assumption}
\label{sec:normality}
In MMAD, the extreme scarcity of anomalies often makes direct discriminative learning unreliable. Hence, many methods adopt the \emph{normality assumption}. Under this assumption, normal multi-modal data exhibit consistent and compact association patterns, and anomalies are identified as deviations from these norms.

\begin{figure*}
    \centering
    \begin{minipage}[t]{0.66\linewidth}
        \centering
        \includegraphics[width=\linewidth]{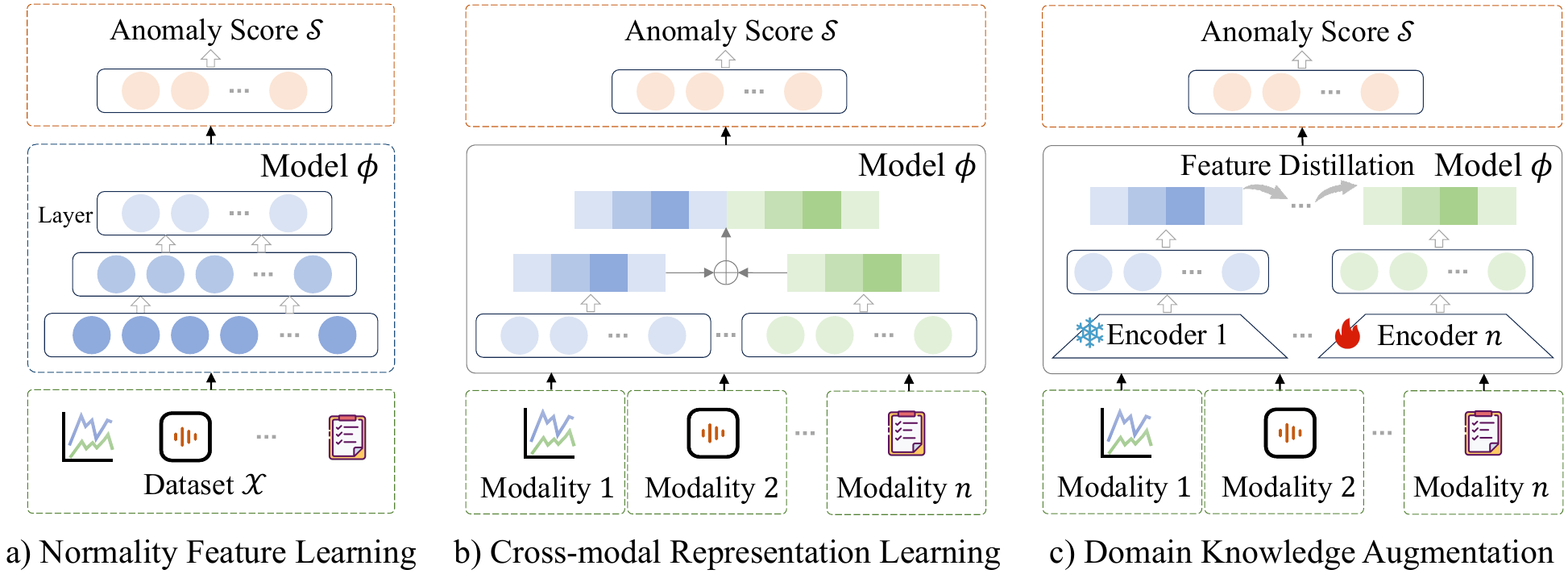}
    \end{minipage}\hfill
    \begin{minipage}[t]{0.32\linewidth}
        \centering
        \includegraphics[width=\linewidth]{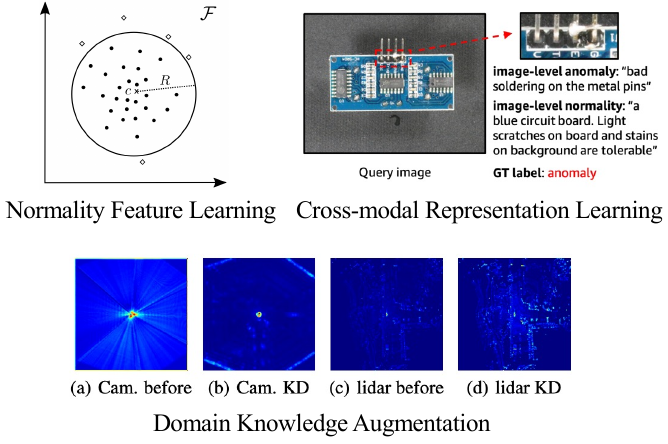}
    \end{minipage}
    \caption{Overview of normality-assumption methods across three groups: a) normal feature learning learns compact normal representations; b) cross-modal representation learning captures inter-modality relations to enrich the normal space; and c) domain knowledge enhancement leverages priors to refine the normal boundary. \revise{The right panel shows schematic or qualitative results adapted from Deep SVDD~\cite{ruff2018deep}, WinCLIP~\cite{jeong2023winclip}, and VeXKD~\cite{ji2024vexkd}.}}
    \label{fig:NS}
\end{figure*}

\revise{
Normality-based methods typically learn a representation $z$ that captures both modality-specific regularities and cross-modal consistency. The scoring function $f$ is often a two-stage mapping with an encoder $\phi$ and a scoring head $\varsigma$, i.e., $f = \varsigma\circ\phi$. Given a test sample with observed modality set $\mathcal{M}$, the anomaly score is computed as
\begin{equation}
    \mathcal{S} = f(\mathcal{X},\mathcal{M}) = \varsigma\big(\phi(\mathcal{X}_\mathcal{M})\big).
\end{equation}
Here, $\varsigma(\cdot)$ can implement different criteria, such as reconstruction error, likelihood under a normal density model, or distance to a compact normal set.
} 
By modeling normal patterns during training, these methods establish a reference for evaluation and flag samples that deviate at test time. As illustrated in Fig.~\ref{fig:NS}, we categorize normality-assumption methods into three groups: normal feature learning, cross-modal representation learning, and domain knowledge enhancement.

\subsection{Normality Feature Learning}

In unsupervised or weakly supervised MMAD, anomaly scarcity and variability make it natural to learn a representation of \emph{normal} patterns from predominantly normal data, thereby characterizing normality and defining the decision boundary (CH2). This often mitigates sensitivity imbalance (CH3) and improves anomaly recall (CH4).

\revise{
Normality feature learning aims to obtain a multimodal embedding $z$ that captures modality-specific regularities and cross-modal consistency. We abstract $\phi(\cdot)$ as a holistic normality encoder that maps a (possibly incomplete) observation to an embedding:
\begin{equation}
    z = \phi\big(\mathcal{X}_{\mathcal{M}}\big).
\end{equation}
In this view, $\phi(\cdot)$ hides architectural choices and can be instantiated via reconstruction, prediction, or density modeling. A key practical issue is \emph{uneven modality maturity} (CH1): strong foundation encoders for vision/text may coexist with shallow, task-specific encoders for bespoke sensors, biasing representation learning; we discuss remedies in Sec.~\ref{subsec:3B}.}
The embedding $z$ thus serves as an implicit normality prior. We group normality feature learning into generic and boundary-constrained methods.

\subsubsection{Generic Normality Feature Learning}
This category focuses on learning features of normal patterns, effectively addressing the challenges arising from the diversity (CH3) and rarity (CH4) of anomalous samples. 
By accurately characterizing the distribution structure of normal data, the model can identify inputs located in low-probability density regions as anomalous, thereby achieving generalized detection of unknown anomalies.

\mypara{Foundations in unimodal settings.}
Inspired by classical data compression and dimensionality reduction techniques, reconstruction-based anomaly detection methods aim to learn compact representations of normal data by enforcing the model to capture essential structural patterns while minimizing reconstruction errors. 
In time-series analysis, LSTM-ED~\cite{hundman2018detecting} proposes an encoder-decoder scheme based on long-short-term memory networks that learns to reconstruct normal time-series behavior.
TranAD~\cite{tuli2022tranad} uses attention-based sequence encoders to swiftly perform inference with the knowledge of the broader temporal trends in the data.
Generative adversarial networks~\cite{goodfellow2020generative} further advanced reconstruction-based detection.
For images domain, f-AnoGAN~\cite{schlegl2019f} builds a generative model of healthy training data.
Subsequent works extended this paradigm to general and industrial images.
GANomaly~\cite{akcay2018ganomaly} uses a conditional generative adversarial network that jointly learns the generation of high-dimensional image space and the inference of latent space.
Another key approach is prediction-based anomaly detection, which uses patterns in previous normal data to predict current data instances to learn feature representations. 
This approach is popular in video anomaly detection, such as Liu et al.~\cite{liu2018future} and AnoPCN~\cite{ye2019anopcn} leverage the difference between a predicted future frame and
its ground truth to detect an abnormal event. 

\mypara{Extensions to multi-modal scenarios.}
Building upon unimodal research, researchers have extended normal feature learning to multi-modal contexts to leverage complementary information from different data sources. 
In this scenario, the research objective shifts from modeling a single distribution to capturing the joint distribution of multiple modalities.
For example, Feng et al.~\cite{feng2023self} train an autoregressive model to generate sequences of audio-visual features, using feature sets that capture the temporal synchronization between video frames and sound. 
Willibald et al.~\cite{willibald2025multimodal} propose a mixture-of-experts framework that integrates the complementary detection mechanisms with a visual-language model for environment monitoring and a Gaussian-mixture regression-based detector for tracking deviations in interaction forces and robot motions. 
MAVAD~\cite{leporowski2024mavad} combines visual and audio features extracted from video sequences by means of cross-attention.

\mypara{Advantages.}
Such methods are relatively simple in structure and exhibit strong generality, making them applicable to various modalities and easily integrable with different neural network architectures for anomaly detection, which enables effective generalization to previously unseen abnormal patterns. 

\mypara{Limitations.}
When the training data are contaminated, the learned feature distributions may inevitably become biased, leading to an inaccurate characterization of normal patterns. Moreover, methods that primarily rely on modeling normal data distributions often fail to capture fine-grained semantic consistency across different modalities, which can further limit discriminative capability and degrade detection performance in multi-modal settings.

\subsubsection{Boundary-constrained Feature Learning}
The paradigm introduces explicit anomaly measures to construct a tight decision boundary that encloses normal samples in the feature space, thereby directly alleviating the issue of ambiguous decision boundaries (CH2).

\mypara{Foundations in unimodal settings.}
Classical one-class methods (e.g., OC-SVM~\cite{scholkopf1999support} and SVDD~\cite{tax1999support}) build explicit boundaries---a hyperplane or minimal-volume hypersphere---to separate normal data from anomalies in the input or kernel space. Deep extensions learn more expressive latent boundaries, exemplified by Deep SVDD~\cite{ruff2018deep} and its variants such as Deep SAD~\cite{ruff2019deep}. For time series, COCA~\cite{wang2023deep} and AOC~\cite{mou2023deep} adapt hypersphere-based objectives with temporal encoders, while RoCA~\cite{mou2025roca} and FOCA~\cite{ma2025foca} further improve robustness under contamination and foundation backbones.

\mypara{Extensions to multi-modal scenarios.}
In multi-modal scenarios, boundary-constrained feature learning incorporates explicit decision boundaries—e.g., hyperspheres or ellipsoids—over fused representations to model normality and mitigate distributional discrepancies jointly.
An early spectral framework~\cite{gao2011spectral} uses tensor or graph embeddings to enclose consistent multi-source patterns within a compact low-rank manifold. 
\revise{Crucially, optimizing such compactness in high-dimensional multimodal spaces risks \emph{representation collapse}, where models may converge to trivial mappings (e.g., collapsing all inputs to a constant point) to trivially minimize the boundary loss. To avoid this, recent extensions explicitly integrate structural regularizations.
For instance, MCDSVDD~\cite{perez2023multi} trains a neural network to map different normal classes to separate hyperspheres in the latent space, maintaining spatial variance rather than collapsing to a single center. 
This method has been successfully applied to astronomical data with multiple object types.
DEMS-SVDD~\cite{wang2025distribution} constructs SVDD boundaries in multi-modal subspaces, combining distribution-entropy and graph regularization to prevent dimensional collapse and handle heterogeneous modality dimensionalities and distributions.
Similarly, MSVDD~\cite{blanco2025mathematical} captures complementary structures across modalities by optimizing multiple Euclidean hyperspheres, and supports kernel extensions to model nonlinear boundaries while avoiding degenerate solutions.}

\mypara{Advantages.}
Boundary-constrained methods provide explicit, interpretable decision boundaries, directly mitigating ambiguity and offering distance-based anomaly scores. 
Focusing solely on normal compactness, they handle anomaly diversity and scarcity effectively, which avoids over-generalization by enforcing the building boundary. 

\mypara{Limitations.}
The compactness assumption can break down for heterogeneous or non-stationary data, yielding loose or unstable decision boundaries.
\revise{Moreover, multimodal boundary objectives are still sensitive to encoder capacity gaps, which can cause \emph{modality dominance} and \emph{collapse} in weaker branches. Thus, compactness constraints alone often require explicit alignment and robust fusion (Sec.~\ref{subsec:3B}).}

\subsection{Cross-modal Representation Learning}
\label{subsec:3B}
In MMAD, cross-modal heterogeneity can place modality-specific features in disparate spaces, worsening modal discrepancy (CH1) and causing uneven sensitivity across modalities (CH3). \revise{In many cases, CH3 is driven by granularity gaps, e.g., aligning dense visual patches with coarse global text or sparse sensor streams.}
Assuming that normal samples are cross-modally consistent while anomalies induce conflicts, cross-modal representation learning detects anomalies by modeling inter-modality relations and scoring deviations from the learned consistency.

\revise{
These approaches typically encode each observed modality and then fuse or interact in the representation space:
\begin{equation}
    z = \Theta\Big(\{\phi_k(\mathbf{x}^{(k)})\}_{k\in\mathcal{M}}\,;\,\mathcal{M}\Big),
\end{equation}
where $\phi_k(\cdot)$ denotes the encoder for modality $k$ and $\Theta(\cdot)$ is an interaction operator (e.g., aggregation, attention, message passing, or shared-space projection). The resulting $z$ is scored by cross-modal disagreement or distance to a learned normal region.
}
We broadly group existing methods into two families: cross-modal semantic synchronization and local consistency verification.

\subsubsection{Cross-modal Semantic Synchronization}
This paradigm aims to synchronize high-level global semantics by mapping features from heterogeneous modalities into a shared semantic space, thereby achieving high-level consistency across modalities (CH1). 

\mypara{Cross-modal fusion strategies.}
\revise{Fusion-based methods instantiate the interaction operator $\Theta(\cdot)$ by combining modalities in three canonical ways~\cite{zhao2024deep}. (1) \textit{Early fusion} aggregates raw inputs before encoding, e.g.,
$z = \phi(\text{Concat}(\{\mathbf{x}^{(k)}\}_{k \in \mathcal{M}}))$.
It preserves fine-grained correlations~\cite{tao2025g2sf} but is most sensitive to modal discrepancy (CH1) (e.g., temporal asynchrony and heterogeneous dimensionality). (2) \textit{Late fusion} first computes modality-specific decisions and then aggregates them at the score level, e.g.,
$S = \Theta_{\text{score}}(\{f_k(\phi_k(\mathbf{x}^{(k)}))\}_{k \in \mathcal{M}})$.
This design can reduce modality dominance in multi-sensor settings~\cite{li2025multi}, but it weakens early semantic interaction and may miss anomalies expressed primarily as cross-modal inconsistency. (3) \textit{Hybrid/Intermediate fusion} balances the above extremes by introducing interactions across intermediate layers, e.g.,
$z^{(l)} = \Theta^{(l)}(\{z_k^{(l-1)}\}_{k \in \mathcal{M}})$.
Although more computationally demanding, it is often most effective for complex inter-modal dependencies in MMAD; representative methods include M3DM~\cite{wang2023multimodal}, adaptive unbalanced fusion designs~\cite{ghadiya2024cross, cheng2025multimodal}, and attention-guided 2D/3D decoders~\cite{ali20252d_3d}.
}


\mypara{Cross-modal alignment.}
Cross-modal alignment reduces modality discrepancy by mapping heterogeneous representations into a shared semantic space via contrastive objectives or feature mapping, often building on CLIP-style vision--language pretraining~\cite{radford2021learning}. By bridging representation spaces without explicit feature fusion, anomalies can be exposed as semantic misalignment or low agreement between modalities. Prompt-based adaptation further tailors the aligned space to downstream localization and detection, as in VadCLIP, AdaCLIP, and AA-CLIP~\cite{wu2024vadclip, cao2024adaclip, ma2025aa}.
\revise{To alleviate sensitivity imbalance from anomaly diversity (CH3), MMAD increasingly uses hierarchical alignment to address \emph{granularity imbalance}: \emph{global} clip--text matching captures macro semantic deviations, while \emph{token-level} patch--word alignment reveals subtle localized defects.}
Beyond CLIP-style alignment, related designs include correspondence learning via cross-modal feature mapping~\cite{costanzino2024multimodal}, decoupled contrastive semantic alignment~\cite{yin2026learning}, and entropy-regularized alignment/optimization that sharpens normal structures~\cite{zeng2024scalable, zeng2025hierarchical}.

\revise{However, \emph{uneven modality maturity} can induce \emph{representation collapse}: with large capacity gaps (e.g., a foundation vision backbone vs. a shallow sensor encoder), contrastive training such as  InfoNCE may push the weaker modality toward trivial alignment and suppress anomaly-relevant cues. Mitigations include asymmetric optimization (e.g., freezing the mature encoder and adapting the weaker branch via lightweight adapters) and auxiliary self-supervised reconstruction to preserve the weaker modality’s structure.}

\mypara{Advantages.}
This paradigm bridges modality heterogeneity by synchronizing high-level global semantics and effectively leveraging complementary information across modalities to build robust normal representations. 
As a result, it generally improves the model’s ability to distinguish diverse types of anomalies.

\mypara{Limitations.}
This paradigm relies on strong cross-modal consistency among normal samples. \revise{Although effective at bridging high-level heterogeneity (CH1), global synchronization can smooth out local details and reduce sensitivity to fine-grained anomalies (CH3), especially when violations occupy only a small fraction of the input.} In practice, weak/noisy relations or missing/imbalanced modalities can cause alignment and fusion to break down, limiting generalization.
\revise{Deployment is also challenging: dual-stream foundation backbones are compute- and memory-intensive, and transformer-based synchronization often scales as $\mathcal{O}(N^2)$ with sequence length. Dense fusion further enlarges the feature space and memory footprint, complicating real-time use on resource-constrained edge devices.}

\subsubsection{Local Consistency Verification}
This paradigm establishes precise correspondences or dynamic constraints at the level of image patches, points, or regions. \revise{By shifting from global semantic mapping to token-level verification, these methods directly address the granularity imbalance (CH3) discussed earlier.}
By emphasizing fine-grained verification, these methods become more sensitive to diverse and complex anomalies (CH3), enabling more accurate anomaly localization and improved overall detection performance.

\mypara{Fine-grained correspondence matching.}
This design exploits geometric or structural priors in normal samples to enforce precise local alignments and reveals anomalies through matching failures or high correspondence errors.
Such methods are commonly used in multi-modal scenarios with strong spatial correspondences, such as RGB–depth, image–point cloud, and 3D industrial inspection settings.
Typical approaches include FiLo++~\cite{gu2025filo++}, a zero-/few-shot method that fuses fine-grained descriptions with deformable localization for patch-level cross-modal matching in industrial and medical anomaly detection.
MPN~\cite{zhao2023patch}, CPMF~\cite{cao2024complementary} and Bergmann et al.~\cite{bergmann2023anomaly} enable robust point-to-point matching and local consistency verification in 3D point clouds.
In RGB-depth scenarios, DADA~\cite{zavrtanik2024cheating} enables learning a general discrete latent space that jointly models RGB and 3D data for 3D surface anomaly detection.
M3DM-NR ~\cite{wang2025m3dm} incorporates patch-level correspondence verification between noisy RGB and depth modalities to resist interference and capture fine-grained geometric discrepancies.

\mypara{Attention-guided consistency constraint.}
Attention-guided consistency constraint methods dynamically enforce local consistency through cross-modal attention mechanisms or regularization.
This allows adaptive discovery of semantically related regions without strict geometric priors.
Such methods often leverage cross-attention to capture inter-modal dependencies and constrain attention maps or activations to remain consistent across modalities.
Recent works such as Patel et al.~\cite{patel2022cross}, MOTCat~\cite{xu2023multimodal}, Gu et al.~\cite{gu2025multi}, and Su et al.~\cite{su2025semantic} introduce cross-modal attention to align semantic features for guiding local consistency.
PathoGraph~\cite{wu2025anomaly} leverages structured clinical records, temporally evolving symptom graphs, and medical ontologies to build semantically interpretable latent spaces.
ASMFD~\cite{fan2024cross} develops cross-modal consistency regularization with aesthetic similarity constraints.

\mypara{Advantages.}
Local Consistency Verification improves MMAD by providing fine-grained sensitivity to subtle local deviations that are often missed by global semantic synchronization, thereby enhancing anomaly localization and overall detection accuracy. \revise{By aligning modalities at the finest possible granularity (e.g., patch-to-token), it prevents high-density information in one modality from being suppressed by lower-density signals in another, effectively mitigating the granularity mismatch in CH3.}

\mypara{Limitations.}
It relies heavily on reliable local correspondences or stable attention patterns in normal samples. When modalities are strongly misaligned, noisy, or partially missing, these assumptions may no longer hold, which can lead to reduced performance in real-world heterogeneous data. 
In addition, such methods often introduce high computational cost due to fine-grained matching or attention operations, which limits their scalability to high-resolution or large-scale multi-modal inputs.

\subsection{Domain Knowledge Augmentation}
To address the limitations of knowledge deficit (CH2) and concept drift (CH5), this approach incorporates structured knowledge priors into the normality assumption framework, thereby constructing more compact and robust representations of normal patterns. 
\revise{By introducing knowledge constraints, these methods guide models to capture the underlying semantic logic of normality in complex and heterogeneous multi-modal data, as follows:
\begin{equation}
z = \Psi(\{\phi_k(\mathbf{x}^{(k)})\}_{k \in \mathcal{M}} \mid \mathcal{K})
\end{equation}
where $\Theta$ denotes the approach exploiting konwledge, $\mathcal{K}$ denotes external knowledge or semantic priors. }
According to the manner in which domain knowledge is introduced and exploited, existing approaches can be broadly categorized into two types: cross-modal knowledge distillation and Prototype Memory Mechanism approaches.

\subsubsection{Cross-modal Knowledge Distillation}
This paradigm aims to transfer the discriminative and semantic knowledge contained in a more informative modality to the target modality or a lightweight student model, thereby injecting domain priors, enhancing the model's ability to characterize the discrimination boundary (CH2), and improving its robustness in concept drift (CH5).

\mypara{Logit-level knowledge distillation.}
Logit-level knowledge distillation transfers softmax logits or soft probabilities from a teacher model trained with richer multi-modal inputs to a student model operating on a target or single modality, enabling the student to inherit inter-class relationships and refine decision boundaries.
Early work, such as LCKD~\cite{wang2023learnable} and MMRD~\cite{gu2024rethinking} introduce distillation paradigm by teacher logits knowledge to handle missing modalities during both training and inference.
$\rm C^2KD$~\cite{huo2024c2kd} employs bidirectional distillation with dynamic on-the-fly selection to generate customized soft labels, effectively bridging the modality gap.
CorrKD~\cite{li2024correlation} proposes a correlation-decoupled schema that models inter-sample, inter-category, and inter-response correlations in logits.
RichKD~\cite{mansourian2025enriching} fuses logits from a dataset-specific unimodal teacher with cross-modal predictions from a pre-trained CLIP model, enriching supervisory signals with semantic diversity.

\mypara{Feature-level knowledge distillation.}
Feature-level knowledge distillation transfers cross-modal structural and semantic information by aligning the intermediate feature representations of teacher and student models. 
Typical approaches include VeXKD~\cite{ji2024vexkd}, TRD~\cite{liu2024multimodal}, A2RD~\cite{chen2025enhancing}, and CMDIAD~\cite{sui2025incomplete}, which employ cross-modal reverse distillation to enhance inter-modality interactions, aiming to selectively transfer beneficial spatial features to single-modal students.
FD-CMKD~\cite{liu2025distilling} disentangles modality-generic and specific information in the frequency domain to facilitate balanced knowledge transfer across heterogeneous modalities.
MUKDF~\cite{wang2026modality} incorporates modality-uncertainty-aware mechanisms with dual-branch extraction and counterfactual reasoning for robust feature distillation in multi-modal sentiment analysis under missing modalities.

\mypara{Advantages.}
Cross-modal knowledge distillation injects domain priors by transferring discriminative and semantic information from informative modalities to target models, improving the characterization of normality decision boundaries.
Logit-level methods are lightweight and easy to integrate, while feature-level variants provide deeper structural knowledge, often yielding state-of-the-art results.

\mypara{Limitations.}
Distillation performance strongly depends on teacher quality and training-time modality availability, which may hinder robustness in highly dynamic multi-modal environments.
The inherent modality heterogeneity can lead to information loss, misalignment, or diluted anomaly signals in fused teacher representations.

\subsubsection{Prototype Memory Mechanism}
By maintaining a prototype memory consisting only of normal patterns as a domain prior, this paradigm constrains the discrimination boundary (CH2), which assumes that the feature distribution of normal samples can be aggregated into a finite number of representative prototypes in the embedding space.

\mypara{Memory bank with prototype clustering.}
This paradigm constructs a compact representation of normality by extracting feature embeddings from training samples and applying clustering algorithms to derive a finite set of representative prototypes. 
Typical methods include CIF~\cite{lin2025commonality} and CPIR~\cite{shangguan2025cpir}, which employ memory to perform prototype clustering that models high-order inter-modal relationships, effectively capturing structural commonality in multi-modal anomaly detection.
MTRMB~\cite{zhou2025multimodal} constructs a multi-modal task representation memory bank via key-prompt-guided cross-modal feature interaction and clustering. 
Information-theoretic extensions further refine prototype quality by minimizing structural entropy to capture intrinsic relational organization in normal data, as demonstrated in hypergraph-based recommendation pre-training~\cite{zhang2025enhanced} and hierarchical text classification optimization~\cite{liu2025hierarchical}.

\mypara{Dynamic memory mechanisms.}
Dynamic memory mechanisms adaptively refine or update stored prototypes and memory entries to accommodate intra-class variations, concept drift, and streaming data, enabling continuous refinement of normality representations without requiring full model retraining.
Recent work, such as DPU~\cite{li2025dpu}, which introduces a plug-and-play framework for multi-modal out-of-distribution detection that dynamically adjusts class-specific prototypes based on sample variance and multi-modal prediction discrepancies.
DNPR~\cite{li5786608dnpr} proposes dynamic normal prototype refinement integrated with progressive masked geometric registration and vision-language models.
FastRef~\cite{tian2025fastref} develops an efficient test-time prototype refinement framework for few-shot industrial anomaly detection, iteratively optimizing transport probabilities for anomaly suppression and transformation matrices for characteristic transfer from query images.
EG-MPC~\cite{hu2025noise} enables high-quality frame prediction and reconstruction under background interference by adaptively refining normality patterns across scales in multi-modal video streams.

\mypara{Advantages.}
This paradigm effectively incorporates multi-modal domain knowledge through clustering or dynamic updates, improving robustness to rare anomalies and unbalanced detection.

\mypara{Limitations.}
Despite their effectiveness, prototype-based memory mechanisms face a critical trade-off between memory capacity and representational diversity. Insufficient capacity limits the coverage of normal variations, whereas excessive capacity increases the risk of memorizing anomalous samples or noise, thereby degrading the purity of normality modeling. 

\section{Methods based on Abnormality Assumption}
\label{sec:anomaly}

The abnormality assumption complements the normality view by explicitly constructing the ``flip side'' of the data distribution. Rather than only reconstructing or enforcing self-consistency on normal data, it leverages domain knowledge to synthesize contrastive signals that expose deviations.
\revise{This paradigm differs from standard self-supervised learning: while both use transformations (masking, perturbation, permutation), self-supervision learns \emph{invariant} representations, whereas pseudo-anomaly generation creates \emph{hard negatives} to make detectors \emph{sensitive} to abnormality.}

\revise{Given a normal observation $\mathcal{X}_{\mathcal{M}}$, an injected sample is generated as $\tilde{\mathcal{X}}_{\mathcal{M}} = T(\mathcal{X}_{\mathcal{M}})$. The detector enforces $f(\mathcal{X}_{\mathcal{M}}) < f(\tilde{\mathcal{X}}_{\mathcal{M}})$ by minimizing
\begin{equation}
\mathcal{L}_{\text{AA}}=\mathbb{E}_{\mathcal{X}_{\mathcal{M}}\sim\mathcal{D}_n}\Big[\ell\big(f(\mathcal{X}_{\mathcal{M}}),\,f(\tilde{\mathcal{X}}_{\mathcal{M}})\big)\Big].
\label{eq:anomaly_assumption}
\end{equation}
Here $\ell(\cdot,\cdot)$ can be a ranking loss or a binary classification loss treating injected samples as pseudo anomalies.}
Injected samples are not meant to match the true anomaly distribution; they probe outside normal support to tighten the boundary.

\begin{figure*}[ht]
  \centering
  \begin{minipage}[t]{0.66\linewidth}
    \centering
    \includegraphics[width=\linewidth]{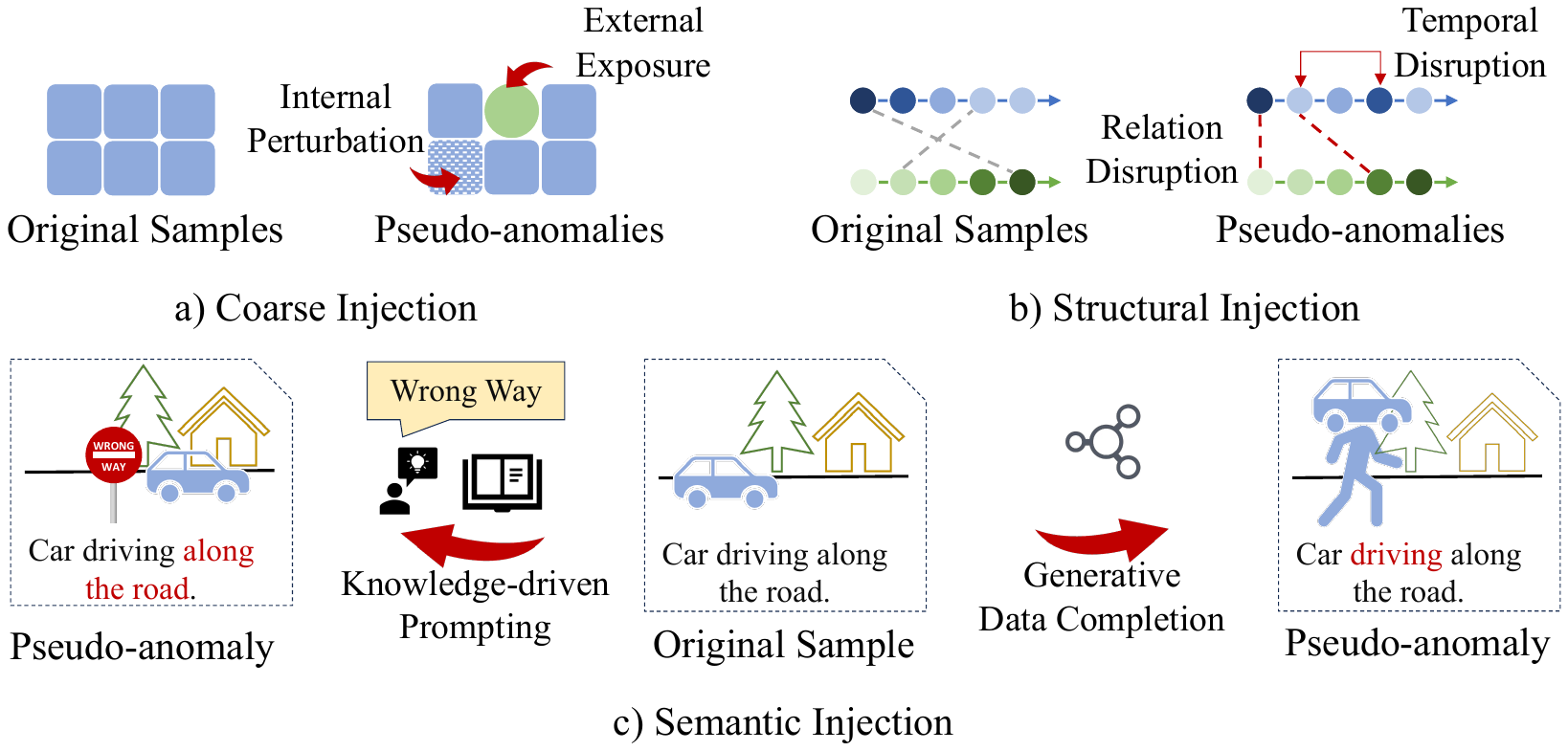}
  \end{minipage}\hfill
  \begin{minipage}[t]{0.30\linewidth}
    \centering
    \includegraphics[width=\linewidth]{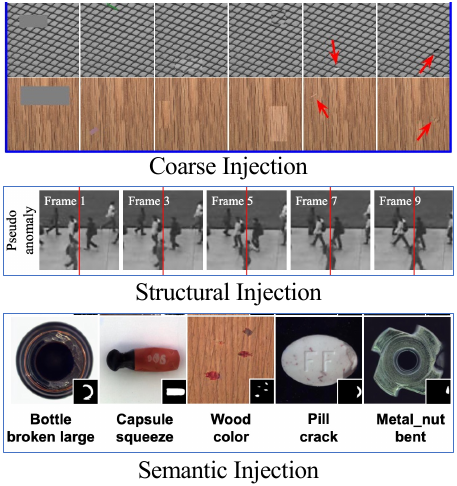}
  \end{minipage}
  \caption{Overview of abnormality-assumption methods that inject pseudo anomalies at three granularities: a) \emph{coarse-grained injection} corrupts low-level statistics via within-modality perturbations or outlier exposure; b) \emph{structural injection} breaks relational integrity by disturbing links or temporal order within/across modalities; and c) \emph{semantic injection} synthesizes high-level logical inconsistencies using external knowledge or generative models. \revise{The right panel shows illustrative pseudo-anomaly visualizations adapted from CutPaste~\cite{li2021cutpaste}, STEAL Net~\cite{astrid2021synthetic}, and AnomalyDiffusion~\cite{hu2024anomalydiffusion}.}}
  \label{fig:AA}
\end{figure*}

We review abnormality-assumption methods in three levels (Fig.~\ref{fig:AA}): \emph{coarse-grained injection} (physical intuition), \emph{structural injection} (relational disruption), and \emph{semantic injection} (knowledge-guided abnormality), progressing from heuristic to semantically grounded boundary learning.
\revise{A shared bottleneck is validating the \emph{rationality} of pseudo-anomalies, which becomes harder from coarse to semantic injection. Coarse/structural schemes typically enforce heuristic or physical constraints (e.g., CutPaste defects within the foreground), whereas semantic completion needs different criteria. Under extreme scarcity in CH4, metrics such as FID may overfit, motivating reference-free measures (e.g., Inception Score, IC-LPIPS~\cite{hu2024anomalydiffusion}) with strict alignment checks; for time series/video, rationality must also respect physical constraints and temporal inertia.}

\subsection{Coarse-grained Injection}
Relying solely on reconstruction of normal data can lead to a loose decision boundary, as discussed in CH4. Coarse-grained injection offers a direct way to ``complete the flip side'' of the data distribution by exposing the detector to low-level corruptions or externally sourced outliers.
By contrasting clean samples with their corrupted counterparts, these methods promote a sharper separation between normal and abnormal regions.
Crucially, coarse-grained injection largely bypasses data heterogeneity in CH1: simple operations such as masking, cutout, and noise injection are modality-agnostic and can be applied to images, time series, and text without semantic alignment.
\revise{
Conceptually, the transformation operator $T(\cdot)$ covers two complementary branches. The first is \emph{internal perturbation}, which applies unstructured, data-level augmentations within each observed modality, namely $T(\mathcal{X}_{\mathcal{M}})=\{\mathcal{A}(\mathbf{x}^{(k)})\}_{k\in\mathcal{M}}$, where $\mathcal{A}(\cdot)$ may instantiate additive Gaussian noise $\mathbf{x}^{(k)}+\boldsymbol{\epsilon}$, Mixup, or adversarial perturbations. The second is \emph{outlier exposure}, which samples proxy anomalies from an auxiliary external dataset, namely $\tilde{\mathcal{X}}_{\mathcal{M}}^{\text{out}}\sim\mathcal{D}_{\text{out}}$. Together, these injections provide coarse contrasts that regularize the boundary and improve robustness to unseen outliers.
}

\subsubsection{External Exposure}
A direct way to address CH4 is to introduce proxy anomalies from external sources. This subclass follows an open-world intuition: any data lying outside the training distribution can be treated as anomalous.

\mypara{Foundations in unimodal settings.} In the CV domain, Outlier Exposure (OE)~\cite{hendrycks2018deep} shows that training detectors with an auxiliary outlier dataset improves generalization to unseen anomalies. Exposure to many ``foreign" examples effectively compacts the decision boundary around normal data. Similarly, Deep SAD~\cite{ruff2019deep} incorporates a small number of external samples, indicating that even limited auxiliary supervision can substantially increase latent-space separation.

Building on OE, subsequent studies extended external exposure to other unimodal domains before moving to multi-modal scenarios.
Latent Outlier Exposure (LOE)~\cite{qiu2022latent} revisits this paradigm under contaminated training data, using auxiliary outliers as latent-space regularizers to prevent representation collapse; it remains effective in other contaminated settings~\cite{mou2025roca}.
Domain-specific variants further adapt the idea to time series~\cite{feng2021unsupervised}, robotic vision~\cite{mantegazza2022outlier}, and anomalous sound detection~\cite{zhang2023outlier} by introducing exogenous temporal patterns, visually irrelevant scenes, or auxiliary audio as proxy anomalies.
In graph-structured data, HGOE~\cite{junwei2024hgoe} combines cross-domain graph outliers with boundary-aware exposure losses in a model-agnostic manner, and has inspired multi-modal adaptations.

\mypara{Extensions to multi-modal scenarios.} As anomaly detection increasingly involves multi-modal data, external exposure has been extended to cross-modal settings.
Recent works~\cite{liu2025extremely} show that simple multi-modal outliers, including cross-modal mismatches and random pairs, can serve as effective proxy anomalies for both detection and segmentation.
These approaches leverage inter-modality misalignment as a strong negative signal, avoiding costly outlier collection and domain-specific annotation.
To mitigate representation bias when the auxiliary exposure set is noisy or partially in-distribution, Taylor Outlier Exposure~\cite{fukuda2024taylor} proposes a Taylor-expanded loss to selectively downweight unreliable samples, addressing cross-modal contamination in MMAD.
More broadly, the survey in~\cite{shi2024outlier} summarizes extensive OE-based methods in visual inspection, while emerging multi-modal benchmarks increasingly adopt exposure-style regularization to better align representations across inputs.

\mypara{Advantages.}
External exposure provides a direct, scalable, and largely modality-agnostic remedy for anomaly scarcity, and it integrates easily with a wide range of detectors and inputs. In multi-modal settings, it supplies a unified negative signal that encourages cross-modal alignment and improves generalization to unseen anomalies without requiring task-specific annotations.

\mypara{Limitations.}
External exposure remains coarse-grained, offering only weak, exclusion-based supervision that may miss complex deviations or subtle cross-modal mismatches. Its effectiveness is also sensitive to exposure quality: poorly aligned or contaminated auxiliary sets can distort the shared boundary.

\subsubsection{Internal Perturbation}
This subclass follows the physical intuition that ``corrupted is anomalous.'' Rather than relying on external data, internal perturbation synthesizes pseudo-anomalies by locally disrupting continuity or statistics in the observed data. It is particularly effective at bypassing data heterogeneity in CH1, since operations such as masking and shuffling are modality-agnostic primitives.

\mypara{Foundations in unimodal settings.}
Internal perturbation originates from data augmentation, later repurposed to generate pseudo anomalies for unsupervised detection, most prominently in vision. Early masking-based methods (e.g., Hide-and-Seek~\cite{kumar2017hide}, Cutout~\cite{devries2017improved}, GridMask~\cite{chen2020gridmask}, Random Erasing~\cite{zhong2020random}) remove or corrupt local regions and can serve as effective pseudo anomalies.
More realistic defects are created via patch manipulation (e.g., CutPaste~\cite{li2021cutpaste}), while interpolative perturbations such as Mixup~\cite{zhang2017mixup} broaden the boundary in input/feature space and have been validated beyond vision, including physiological and time-series settings~\cite{guo2023empirical, carmona2021neural}.

\mypara{Extensions to multi-modal scenarios.}
Internal perturbation extends naturally to MMAD because it relies on simple, modality-agnostic operations and requires minimal domain knowledge. Although fewer works exist than in unimodal vision, recent studies have started adapting masking and perturbation strategies to heterogeneous inputs.
For example, SeMAnD~\cite{reshetova2023semand} applies self-supervised masking to paired vector geometries and imagery, creating pseudo-anomalies by corrupting one modality while maintaining cross-modal consistency checks. Related designs use random masking or feature-level misalignment to exploit inter-modality inconsistencies as proxy signals.
In industrial and security contexts, autoencoder-based fusion models follow the same intuition. Khan et al.~\cite{khan2025robust} apply reconstruction perturbations across sensors for real-time vehicle-damage detection, while integrated deep learning autoencoders~\cite{syed2025advanced} use coarse internal corruptions on diverse streams for cloud anomaly classification. Extensions of CutPaste-style operations to multi-modal fusion and adversarial internal augmentations~\cite{benabderrahmane2025adversarial} further show promise for industrial and cyber-physical systems.

\mypara{Advantages.}
Internal perturbation offers a simple, computationally efficient, and highly modality-agnostic remedy for anomaly scarcity by generating diverse pseudo-anomalies from in-distribution data. These coarse heuristics capture local deviations effectively and avoid the domain-shift risks of external exposure; in MMAD, they can also encourage cross-modal consistency and robustness.

\mypara{Limitations.}
Internal perturbation remains heuristic and coarse-grained, and may miss subtle semantic or structurally complex anomalies, particularly when fine-grained inter-modality dependencies matter. Performance can be sensitive to hyperparameters, and unrealistic artifacts may hurt generalization.

\subsection{Structural Injection}
Structural injection extends coarse-grained perturbations under the intuition that \emph{disrupted structure is anomalous}. It synthesizes pseudo-anomalies by violating temporal or relational regularities of normal data, enabling multi-scale disruptions (from local subsequences to global relations) to address granularity imbalance (CH3) and improve robustness under drift (CH5).
\revise{Concretely, structural injection perturbs the multi-modal observation \(\mathcal{X}_{\mathcal{M}}\) via a structure-disrupting transformation \(T(\cdot)\). Typical instantiations include: (i) \emph{temporal disruption}, which permutes sequential content within each observed modality, i.e., \(T(\mathcal{X}_{\mathcal{M}}) = \{\mathbf{x}^{(k)}_{\pi}\}_{k \in \mathcal{M}}\) where \(\pi\) denotes a permutation operator (e.g., shuffling or reversal); and (ii) \emph{relational disruption}, which replaces local neighborhoods or links, i.e., \(T(\mathcal{X}_{\mathcal{M}}) = \big\{ \mathbf{M}^{(k)} \odot \mathbf{x}^{(k)} + (1 - \mathbf{M}^{(k)}) \odot \psi(\mathbf{x}^{(k)}) \big\}_{k \in \mathcal{M}}\), where \(\mathbf{M}^{(k)}\) is a relational mask and \(\psi(\cdot)\) performs a local relational replacement (e.g., edge dropping or patch swapping).
These contrasts guide the model to place the decision boundary around regions of structural incoherence, rather than around marginal feature deviations.}

\subsubsection{Temporal Structure Disruption}
This sub-subclass targets the temporal consistency inherent in sequential data, operating on the intuition that ``disordered chronology is anomalous." By explicitly violating the natural order, speed, or predictive continuity of normal sequences, these methods generate pseudo-anomalies that simulate contextual deviations, offering finer granularity than coarse masking while remaining computationally tractable.

\mypara{Foundations in unimodal settings.}
The paradigm of temporal structure disruption originated in unimodal time series and video analysis, where sequential integrity serves as a strong supervisory signal. Early Transformer-based approaches like TranAD~\cite{tuli2022tranad} introduced attention masking and subsequence perturbations to enforce robust temporal encoding.  AnomalyBERT~\cite{jeong2023anomalybert} and RedLamp~\cite{obata2025robust} employ data degradation or transformation that combine masking with explicit order disruption and future prediction failures, creating diverse pseudo-anomalies for self-supervised learning.
CutAddPaste~\cite{wang2024cutaddpaste} further refines this by cutting patches informed by abnormal knowledge and pasting them into normal contexts, deliberately breaking local temporal continuity to mimic realistic deviations.
In video anomaly detection, synthetic temporal perturbations~\cite{astrid2021synthetic, rai2024video} leverage snippet shuffling and reversal as negative augmentations to amplify temporal inconsistencies, while recent reviews~\cite{zhu2024advancing} consolidate these order-based disruptions as foundational techniques for sequential representation learning.

\mypara{Extensions to multi-modal scenarios.}
In multi-modal contexts, temporal disruption becomes particularly powerful by exploiting cross-modal consistency as an additional structural cue. Deliberate misalignment between modalities including such as audio-video temporal offsets or sensor stream desynchronization, generates highly informative pseudo-anomalies that reflect real-world multi-modal incoherence.
Fusion frameworks~\cite{wang2025multimodal} introduce controlled audio-video temporal misalignments during training to strengthen joint representations in complex environments. Similarly, Multi-modal VAD~\cite{wang2025vad} incorporates audio-vision-language streams, using modality-specific temporal shifts as negative signals for intelligent monitoring systems. Recurrent architectures~\cite{tariq2024recurrent} combine spatio-temporal audio-visual features, simulating anomalies via selective shuffling or delay injection across modalities. Applications in crowd analysis~\cite{dionis2024multimodal} further demonstrate the efficacy of temporal inconsistencies between visual trajectories and contextual cues in open-access multi-modal setups.

\mypara{Advantages.}
Temporal structure disruption provides finer-grained supervision than coarse perturbations, effectively capturing contextual and sequential anomalies (CH3) while enhancing robustness to distribution shifts (CH5) through structure-aware regularization. In multi-modal settings, cross-modal misalignment offers a semantically rich negative signal, promoting synchronized representations and improved generalization to real inter-modal deviations.

\mypara{Limitations.}
Despite its advantages, reliance on explicit temporal assumptions can limit applicability to non-sequential data, and aggressive disruptions risk introducing artifacts that deviate from realistic anomalies. Sensitivity to disruption strength (e.g., shuffle intensity or offset magnitude) remains a challenge, particularly in multi-modal scenarios where subtle misalignments dominate but are harder to calibrate consistently.

\subsubsection{Relational Structure Disruption}
This sub-subclass emphasizes the relational topology inherent in interconnected data, operating on the intuition that ``fractured connections are anomalous." By deliberately perturbing the dependencies among entities, such as through edge deletions, relation shuffling, or subgraph masking, these methods generate pseudo-anomalies that simulate network-level inconsistencies, addressing granularity imbalance (CH3) by capturing anomalies at varying relational scales while countering concept drift (CH5) via topology-robust regularizations.

\mypara{Foundations in unimodal settings.}
Relational structure disruption stems from graph anomaly detection, where self-supervision uses topological perturbations (edge/attribute edits, relation shuffling) as supervisory signals. Representative methods include DAGAD~\cite{liu2022dagad} and related robustness-oriented designs~\cite{zhou2023improving}; in video, spatio-temporal perturbations can be interpreted as breaking structured dependencies~\cite{rai2024video}.
Later work strengthens this idea with contrastive and adversarial perturbations, e.g., multi-scale contrastive views~\cite{duan2023graph}, dual-bootstrapped training~\cite{tang2024dualgad}, adversarial/multiplex edge dropping~\cite{chen2025adedgedrop, li2025umgad}, and unified perturbation frameworks~\cite{xu2025revisiting, song2025uniform}.

\mypara{Extensions to multi-modal scenarios.}
In multi-modal contexts, relational disruption extends to dependencies among modals, such as perturbing alignments between visual graphs and textual relations or fusing point cloud topologies with image features. This creates pseudo-anomalies that exploit relational voids across modalities, like mismatched object predicates or desynchronized sensor graphs.
Zhang et al.~\cite{zhang2025unified} disrupts cross-modal relations via matching cost filtering in point cloud-image pairs. Generative approaches~\cite{qiao2024generative} incorporate relation perturbations in multi-modal node attributes for semi-supervised learning. These techniques demonstrate the adaptability of relational disruption in fusing heterogeneous structures, enhancing detection in scenarios like industrial inspection or autonomous systems.

\mypara{Advantages.}
Relational structure disruption captures intricate inter-entity dependencies at multiple scales (CH3), yielding interpretable pseudo-anomalies that bolster robustness to evolving distributions (CH5). In multi-modal settings, it uncovers and makes use of cross-modal inconsistencies, fostering unified representations that generalize across fused relational spaces.

\mypara{Limitations.}
These methods assume well-defined relational structures, constraining applicability to sparse or unstructured data, and perturbation designs (e.g., drop ratios) may bias toward specific anomaly types. In multi-modal applications, aligning disruptions across modalities introduces calibration challenges, potentially amplifying noise in high-dimensional relational contexts.

\subsection{Semantic Injection}
In many real-world tasks, anomalies are semantic deviations that violate domain knowledge or contextual plausibility (CH2), which low-level or purely structural perturbations often fail to capture. Enabled by generative and multi-modal foundation models, semantic injection creates knowledge-guided contrasts to sharpen anomaly definitions and partially alleviate scarcity (CH4).
\revise{It can further incorporate external knowledge or semantic priors $\mathcal{K}$ to construct stronger, context-violating contrasts. Typical instantiations include (i) \emph{knowledge-driven prompting},
$T(\mathcal{X}_{\mathcal{M}}) = \mathcal{X}_{\mathcal{M}} \oplus f_{\text{LLM}}(\mathcal{K}_{\text{prompt}})$,
where $f_{\text{LLM}}$ generates rule-violating prompts guided by $\mathcal{K}_{\text{prompt}}$; and (ii) \emph{generative data completion},
$T(\mathcal{X}_{\mathcal{M}}) = f_{\text{gen}}(\mathbf{M} \odot \mathcal{X}_{\mathcal{M}}; \mathcal{K}_{\text{gen}})$,
where $f_{\text{gen}}$ hallucinates out-of-distribution content under generative priors $\mathcal{K}_{\text{gen}}$. These contrasts tighten the decision boundary by emphasizing incoherence over marginal feature deviations.
}

\subsubsection{Knowledge-driven Prompting}
This sub-subclass follows the intuition that ``semantically implausible is anomalous.'' \revise{It is increasingly enabled by foundation models, ranging from contrastive vision--language models (VLMs) to generative multi-modal large language models (MLLMs).}
By injecting semantic priors through designed or learned prompts, these methods synthesize knowledge-guided pseudo-anomalies that violate domain-specific logic or contextual plausibility, without explicitly modeling structural relations.

\mypara{Foundations in unimodal settings.}
Knowledge-driven prompting emerged from zero-/few-shot anomaly detection, where VLMs offer a semantic space to define abnormality via text. WinCLIP~\cite{jeong2023winclip} introduced class-specific negative prompts, while AnomalyCLIP~\cite{zhou2023anomalyclip} learns object-agnostic prompts from normal data to reduce manual engineering.
Later works sharpen this semantic boundary via simpler or learned prompting and context cues, including SimCLIP~\cite{deng2024simclip}, PromptAD~\cite{li2024promptad}, and VCP-CLIP~\cite{qu2024vcp}; larger-model variants (e.g., AnomalyGPT~\cite{gu2024anomalygpt}, AA-CLIP~\cite{ma2025aa}) improve reasoning and robustness, with UniAD~\cite{xiang2025uniad} further stabilizing prompts across domains. 

\mypara{Extensions to multi-modal scenarios.}
In multi-modal contexts, knowledge-driven prompting uses cross-modal semantic priors to expose logical inconsistencies, enabling the synthesis of semantically implausible pseudo-anomalies by reasoning over multiple data streams. For example, Negative Prompting~\cite{nie2024out} uses negative descriptions to simulate anomalous patterns in a multi-modal latent space, while prior work~\cite{xu2025towards} prompts large VLMs to surface cross-modal conflicts (e.g., images contradicting their textual descriptions).

\revise{Recently, integrating MLLMs has shifted the focus from semantic alignment to high-level reasoning. VMAD~\cite{deng2025vmad} exploits the spatial–semantic capabilities of MLLMs for location-aware zero-shot detection, and Holmes-VAD~\cite{zhang2024holmes} uses MLLMs to generate unbiased, explainable justifications for video anomalies, moving beyond numerical scores. As highlighted in recent work~\cite{dang2024explainable}, MLLM reasoning chains offer a conceptual roadmap for advancing MMAD from “perception” to “understanding,” where cross-modal logical conflicts serve as explicit evidence of abnormality.}

\mypara{Advantages.}
Knowledge-driven prompting provides semantically rich supervision (CH2) by directly incorporating domain knowledge through VLMs, reducing definition ambiguity and enabling precise anomaly localization. \revise{This paradigm is also more explainable than normality-based reconstruction: whereas reconstruction error is a ``black-box" signal, semantic injection offers explicit contrastive evidence. By comparing inputs to predefined abnormal concepts, these methods yield human-readable justifications, crucial for high-stakes big data applications.} It further mitigates class imbalance (CH4) by generating diverse pseudo-anomalies on demand, supporting zero-/few-shot scenarios and strong cross-domain generalization without task-specific training data.

\mypara{Limitations.}
These methods depend heavily on pre-trained LLM quality, limiting performance in domains with poor semantic coverage or noisy prompts. Prompt engineering is sensitive to design choices, and large models incur high computational costs. In multi-modal settings, maintaining consistent semantic alignment across modalities is especially difficult for rare or emerging anomaly types.

\begin{table*}[t]
\centering
\caption{\revise{Quantitative performance and theoretical taxonomy of representative MMAD methods across diverse domains.}}
\label{tab:performance_comparison}
\renewcommand{\arraystretch}{1.2}
{\footnotesize
\setlength{\tabcolsep}{3pt}
\begin{tabular}{l c l c p{4.8cm} p{2.6cm} p{2.6cm}}
\toprule
\textbf{Method} & \textbf{Modality} & \textbf{Paradigm} & \textbf{FM} & \textbf{Key Mechanism \& Intuition} & \textbf{Representative Benchmark} & \textbf{Performance (Metrics)} \\
\midrule
\multicolumn{7}{c}{\textbf{\textit{Image Anomaly Detection (Industrial, Medical \& General Vision - Metrics: Image/Pixel-AUROC)}}} \\
\midrule
CutPaste~\cite{li2021cutpaste} & I & AA (Coarse) & - & Heuristic patch self-augmentation & MVTec AD & 95.2\% (I-AUROC) \\
TaylorOE~\cite{fukuda2024taylor} & I & AA (Coarse) & - & Polynomial Taylor-expansion OE & CIFAR / ImageNet & 99.3\%+ (AUROC) \\
M3DM~\cite{wang2023multimodal} & I+3D & NA (Alignment) & \checkmark & Point-to-feature semantic alignment & MVTec 3D-AD & 94.5\% (I-AUROC) \\
WinCLIP~\cite{jeong2023winclip} & I+L & AA (Semantic) & \checkmark & Compositional prompt ensembling & MVTec AD / VisA (Zero-shot) & 91.8\% / 78.1\% (AUROC) \\
PromptAD~\cite{li2024promptad} & I+L & AA (Semantic) & \checkmark & Knowledge-guided text descriptions & MVTec AD (Few-shot) & 94.6\% (I-AUROC) 95.9\% (P-AUROC) \\
FastRef~\cite{tian2025fastref} & I & NA (Knowledge) & \checkmark & Dynamic prototype memory distillation & MVTec AD  & 93.8\% (I-AUROC) 95.7\% (P-AUROC) \\
AdaCLIP~\cite{cao2024adaclip} & I+L & NA (Alignment) & \checkmark & Adaptive prompt-based fine-tuning & MVTec AD & 89.2\% (I-AUROC) 88.7\% (P-AUROC)\\
AnomalyDiffusion~\cite{hu2024anomalydiffusion} & I+L & AA (Semantic) & \checkmark & Generative pseudo-anomaly synthesis & MVTec AD / VisA & 99.1\% (P-AUROC) 99.2\% (I-AUROC) \\
MMRD~\cite{gu2024rethinking} & I+3D & NA (Knowledge) & - & Teacher-student knowledge distillation & MVTec 3D-AD & 95.0\% (I-AUROC) 97.6\% (PRO) \\
\midrule
\multicolumn{7}{c}{\textbf{\textit{Video Anomaly Detection (Surveillance \& Events - Metrics: Frame-level AP / AUC)}}} \\
\midrule
VadCLIP~\cite{wu2024vadclip} & V+L & NA (Alignment) & \checkmark & Dual-branch visual-language alignment & XD-Violence & 84.5\% (AP) \\
DSANet~\cite{yin2026learning} & V+L & NA (Alignment) & \checkmark & Semantic-aware spatio-temporal modeling & UCF-Crime / XD-Violence & 89.44\% / 86.95\% (AP) \\
STEAL Net~\cite{astrid2021synthetic} & V & AA (Structural) & - & Breaking temporal regularities & Pre2 & 98.4\% (AUC) \\
\midrule
\multicolumn{7}{c}{\textbf{\textit{Time-Series, Sensors \& IT Systems (Metrics:  F1 / Precision)}}} \\
\midrule
TranAD~\cite{tuli2022tranad} & T & NA (Feature) & - & Attention-based normal reconstruction & SMD  & 96.1\% (PA F1) 99.7\% (AUC) \\
FOCA~\cite{ma2025foca} & T & NA (Feature) & \checkmark & Boundary-constrained FM adaptation & WADI & 74.9\% (PA F1) \\
AnomalyBERT~\cite{jeong2023anomalybert} & T & AA (Structural) & - & Sequence relational masking & SWaT & 85.4\% (PW F1) 92.5\% (PA F1) \\
RedLamp~\cite{obata2025robust} & T & AA (Structural) & - & Classification of perturbations & UCR & 89.7\% (VUS-AUC) \\
CutAddPaste~\cite{wang2024cutaddpaste} & T & AA (Structural) & - & Break decomposition structure  & UCR & 68.2\% (RPA-F1) \\
\bottomrule
\multicolumn{7}{p{\textwidth}}{\footnotesize \textbf{Note:} Modality abbreviations: \textbf{I} (Image), \textbf{V} (Video), \textbf{L} (Language/Text), \textbf{A} (Audio), \textbf{T} (Time-series), \textbf{3D} (Point Cloud). Assumption abbreviations: \textbf{NA} = Normality Assumption, \textbf{AA} = Anomaly Assumption. Performance varies by metric (e.g., Pixel-AUROC for industrial defects vs. Frame-AP for video events).} \\
\bottomrule
\end{tabular}
}
\end{table*}

\subsubsection{Generative Data Completion}
This sub-subclass employs generative architectures to synthesize semantically implausible continuations or completions from normal data, operating on the intuition that ``semantically incoherent synthesis is anomalous." By conditioning generation on normal samples while introducing controlled violations of logical or contextual consistency, these methods produce high-fidelity pseudo-anomalies that capture subtle, knowledge-driven irregularities.

\mypara{Foundations in unimodal settings.}
Generative data completion began with GAN-based methods for synthesizing anomalies in unimodal industrial images. Early works such as Doping~\cite{lim2018doping}, Old is Gold~\cite{zaheer2020old}, G2D~\cite{pourreza2021g2d}, Defect-GAN~\cite{zhang2021defect}, and DSG~\cite{niu2020defect} use GANs to augment training with synthetic defects, improving reconstruction-based detectors by exposing them to generated outliers. Few-shot Defect Generation~\cite{duan2023few} extends this with defect-aware feature manipulation for limited data.
Diffusion models then enabled higher-quality synthesis. Building on DDPM~\cite{ho2020denoising}, AnoDDPM~\cite{wyatt2022anoddpm} employs simplex noise for robust anomaly scoring. Recent diffusion frameworks such as RealNet~\cite{zhang2024realnet}, AnomalyDiffusion~\cite{hu2024anomalydiffusion}, Multi-Class Diffusion~\cite{he2024diffusion}, CAGEN~\cite{jiang2024cagen}, and DCP~\cite{dai2024generating} provide controllable, few-shot anomaly generation with superior fidelity and diversity for unsupervised detection.

\mypara{Extensions to multi-modal scenarios.}
Multi-modal extensions use generative models to synthesize cross-modal inconsistencies, such as implausible combinations of visual, depth, or sensor data. AnomalyXFusion~\cite{hu2024anomalyxfusion} applies diffusion to fuse image-text-mask modalities and create semantically violating samples. MMCD~\cite{flaborea2023multimodal} conditions generation on skeleton motion for video anomaly detection, while MAGE-ID~\cite{loodaricheh2025mage} uses GAN–diffusion hybrids to synthesize minority-class threats in network–sensor data. 
Recent work further integrates LLMs to generate profile-conditioned yet semantically anomalous behaviors from historical user traces, enabling proactive graph evolution and adversarial detection~\cite{zeng2025proactive}.
These methods demonstrate that generative models can produce realistic multi-modal pseudo-anomalies, improving detection in industrial and surveillance settings.

\mypara{Advantages.}
Generative data completion produces high-fidelity, semantically coherent pseudo-anomalies (CH2), enabling precise modeling of complex irregularities via controllable synthesis. It mitigates anomaly rarity (CH4) by generating unlimited diverse samples, supporting few-shot and zero-shot generalization and surpassing heuristic methods in reconstruction quality and localization.

\mypara{Limitations.}
These methods are computationally expensive, especially diffusion models with iterative sampling, limiting real-time use. GANs still face training instability and mode collapse, and multi-modal extensions require careful cross-modal conditioning to avoid artifacts. Generalization to highly diverse or unseen anomalies is limited by the generator’s training exposure. \revise{The computational cost of generative paradigms remains a central bottleneck: diffusion-based methods require iterative denoising with many forward passes, causing extreme inference latency that effectively rules out raw deployment in strict real-time anomaly detection and forces a trade-off between semantic anomaly diversity and detection efficiency.}

\revise{\mypara{Summary and FM perspective.} Abnormality-based modeling progresses from low-level perturbations to structure-aware disruptions and, ultimately, knowledge-driven semantic reasoning, yielding increasingly tight and interpretable decision boundaries through explicit contrastive evidence. Across this spectrum, Foundation models ---including MLLMs---are best viewed as shared enablers: under the \emph{Normality Assumption} (Sec.~\ref{sec:normality}) they serve as universal extractors/aligners that reduce heterogeneity and capacity gaps, whereas under the \emph{Anomaly Assumption} (Sec.~\ref{sec:anomaly}) they act as semantic generators/reasoners that support promptable pseudo-anomaly synthesis and, increasingly, natural-language rationales for violated cross-modal constraints. Table~\ref{tab:performance_comparison} summarizes representative methods from both paradigms and their FM integration; the next section then grounds this taxonomy with representative benchmarks and evaluation settings.}

\section{Applications, Benchmarks, and Evaluation}
\label{sec:datasets}
MMAD mitigates the ambiguity of single-stream observations by integrating heterogeneous sources. Accordingly, this section organizes widely used benchmarks by their physical characteristics, spanning high-precision geometric inspection in industrial systems, life-critical sensing in medical applications, and semantically rich surveillance scenarios. \revise{A comprehensive taxonomy of these benchmarks---grouped by modality composition, spatiotemporal scale, and environmental dynamics---is provided in Table~\ref{tab:all_datasets}.}

\subsection{Application Domains and Benchmarks}
\mypara{Industrial and cyber-physical systems.} Industrial anomaly detection is shifting from localized 2D surface defects to full-view 3D inspection and logical constraint verification, amplifying modal discrepancy (CH1) across heterogeneous 2D/3D representations and weakening the availability of explicit semantic priors (CH2). Representative inspection benchmarks span 2D/3D, depth, and logical settings~\cite{bergmann2019mvtec, zou2022spot, bergmann2022beyond, bergmann2021mvtec, liu2023real3d, bonfiglioli2022eyecandies, mishra2021vt, jezek2021deep, tabernik2020segmentation, wieler2007weakly}. Beyond manufacturing, system software monitoring and cyber-physical process control mix logs with continuous multivariate streams and emphasize long-horizon collective anomalies (CH3) and robustness under non-stationary drift (CH5)~\cite{oliner2007supercomputers, xu2009detecting, goh2016dataset, ahmed2017wadi, su2019robust}.

\begin{table*}[!htb]
\centering
\caption{\revise{Summary of multi-modal and unimodal anomaly detection datasets across diverse application domains.}}
\label{tab:all_datasets}
\renewcommand{\arraystretch}{1.3}
\resizebox{\textwidth}{!}{
\begin{tabular}{l c c l l c l c}
\toprule
\textbf{Dataset} & \textbf{Year} & \textbf{Modality} & \textbf{Target / Highlight} & \textbf{Scale / Size} & \textbf{Anomaly Ratio} & \textbf{Annotation Quality} & \textbf{Dynamics} \\ 
\midrule
\multicolumn{8}{c}{\textbf{\textit{Industrial \& Manufacturing Inspection}}} \\
\midrule
MVTec AD~\cite{bergmann2019mvtec} & 2019 & I & Structural defects (scratches, dents) & 5,354 images & $\sim$23\% & Pixel-level mask & Static \\
MVTec LOCO~\cite{bergmann2022beyond} & 2022 & I & Structural + Logical (rule violations) & 3,644 images & $\sim$16\% & Pixel-level mask & Static \\
MVTec 3D-AD~\cite{bergmann2021mvtec} & 2021 & I+3D & Geometric shapes (holes, protrusions) & 4,147 scans & $\sim$23\% & Pixel \& Voxel-level & Static \\
Real3D-AD~\cite{liu2023real3d} & 2023 & High-res 3D & Largest high-precision 360$^\circ$ dataset & 1,250 3D models &  $\sim$\revise{48\%} & Voxel-level mask & Static \\
VisA~\cite{zou2022spot} & 2022 & I & Complex surfaces, multiple instances & 10,821 images & $\sim$11\% & Pixel-level mask & Static \\
Eyecandies~\cite{bonfiglioli2022eyecandies} & 2022 & I+Depth & High-fidelity synthetic data avoiding bias & 10,000 images &  $\sim$25\% & Pixel \& Depth map & Static \\
BTAD~\cite{mishra2021vt} & 2021 & I & Real-world body and surface defects & 2,830 images & $\sim$11\% & Pixel-level mask & Static \\
MPDD~\cite{jezek2021deep} & 2021 & I & Painted metal parts manufacturing & 1,346 images & $\sim$21\% & Pixel-level mask & Static \\
SDD~\cite{tabernik2020segmentation} & 2020 & I & Surface crack detection (weakly sup.) & 400 images & $\sim$13\% & Pixel-level mask & Static \\
DAGM~\cite{wieler2007weakly} & 2007 & I & Historical reference for optical inspection & 8,050 images & $\sim$12\% & Image-level label & Static \\
HDFS~\cite{xu2009detecting} & 2009 & L+T & Distributed systems monitoring & 24.4M logs & $\sim$0.1\% & Sequence-level & Static \\
BGL~\cite{oliner2007supercomputers} & 2007 & L+T & Supercomputer temporal alignment & 4.7M logs & $\sim$7.3\% & Log-level & Static \\
SWaT~\cite{goh2016dataset} & 2016 & T & Critical infrastructure physical attacks & 11 days & $\sim$11.9\% & Point-level & Static \\
WADI~\cite{ahmed2017wadi} & 2017 & T & Water treatment distribution attacks & 16 days & $\sim$5.9\% & Point-level & \revise{Concept Drift} \\
SMD~\cite{su2019robust} & 2019 & T & Cloud server cluster health & 5 weeks & $\sim$4.1\% & Point-level & Static \\
\midrule
\multicolumn{8}{c}{\textbf{\textit{Medical Diagnostics \& Healthcare}}} \\
\midrule
MMR~\cite{li2025adapting} & 2025 & I+OCT & Retinal artery occlusion (multi-modal) & 370 pairs & 28\% & Case \& Pixel-level mask & Static \\
IDRiD~\cite{porwal2020idrid} & 2020 & I & Diabetic retinopathy micro-lesions & 516 images & Imbalanced &  Pixel \& Image-level & Static \\
HeadCT~\cite{chilamkurthy2018deep} & 2018 & I & Urgent care triage (hemorrhage) & 21,586 scans & $\sim$12\% & Image-level label & Static \\
BrainMRI~\cite{baid2021rsna} & 2021 & I & 3D volume neuro-oncological imaging & 8,160 mpMRI scans & 100\% & 3D Voxel-level & Static \\
Br35H~\cite{tbkk-q937-25} & 2025 & I & Brain tumor presence (binary class.) & 3,060 images & 50\% & Image-level label & Static \\
COVID-19~\cite{chowdhury2020can} & 2020 & I & Viral manifestations in lung imaging & 3,487 images & $\sim$12\% & Image-level label & Static \\
ISIC~\cite{codella2018skin} & 2018 & I & Skin cancer / Melanoma standard & 2,750 images & $\sim$19\% & Pixel \& Image-level \& Label & Static \\
CVC-ClinicDB~\cite{bernal2015wm} & 2015 & I & Colon tumors and polyps on surfaces & 612 frames & 100\% & Pixel-level mask & Static \\
Kvasir / Endo~\cite{jha2019kvasir, hicks2021endotect} & 2019/2021 & I & Multi-class GI tract screening & 8,000 images & $\sim$12\% & Image-level \& Pixel-level & Static \\
TN3k~\cite{gong2021multi} & 2021 & I & Thyroid nodule segmentation priors & 4,510 images & 100\% & Pixel-level mask & Static \\
MIT-BIH~\cite{moody2001impact} & 2001 & T & Foundational bio-electrical series & 48 records (116,000 beats) & $\sim$25\% & Point-level & Static \\
ECG5000~\cite{chen2015general} & 2000 & T & Large-scale heart signal screening & 5,000 seqs & $\sim$42\% & Sequence-level & Static \\
MIMIC-III~\cite{johnson2016mimic} & 2016 & T+L & Critical care decompensation \& risk & 61,532 ICU stays & $\sim$11\% & Patient-level & \revise{Concept Drift} \\
\midrule
\multicolumn{8}{c}{\textbf{\textit{Video \& Surveillance Analytics}}} \\
\midrule
UCF-Crime~\cite{sultani2018real} & 2018 & V+Flow & 13 types of criminal behavior & 1,900 videos & $\sim$50\% & Video \& Temporal-level label & Static \\
XD-Violence~\cite{wu2020not} & 2020 & V+A & Largest violence dataset with audio & 4,754 clips & $\sim$41\% & Video \& Frame-level label & Static \\
ShanghaiTech~\cite{luo2017revisit} & 2017 & V & Multi-scene campus surveillance & 316,154 frames & $\sim$13\% & Pixel\& Frame-level lebel & Static \\
VAAR~\cite{ali2025avar} & 2025 & V+A & Synchronized audio-visual cues & 3,000 videos & $\sim$90\% & Video-level label & Static \\
MSAD~\cite{zhu2024advancing} & 2024 & V & Large-scale multi-scenario (14 envs) & 1,010,480 frames & $\sim$31\% & Frame \& Video-level label & Static \\
UCF-Crime-DVS~\cite{qian2025ucf} & 2025 & V & First large-scale DVS benchmark & 1,900 clips & $\sim$48\% & Video \& Temporal-level label & Static \\
\bottomrule
\multicolumn{8}{p{1.25\textwidth}}{\footnotesize \textbf{Modality abbreviations:} \textbf{I} (Image), \textbf{V} (Video), \textbf{L} (Language/Text), \textbf{A} (Audio), \textbf{T} (Time-series), \textbf{3D} (Point Cloud), \textbf{Depth} (Depth Map), \textbf{OCT} (Optical Coherence Tomography).
\textbf{Annotation Quality:} Defines the granularity of evaluation.
\textbf{Dynamics:} Indicates whether the dataset assumes a stationary environment ('Static') or contains evolving conditions and non-stationary shifts ('Concept Drift').} \\
\bottomrule
\end{tabular}
}
\end{table*}

\mypara{Medical diagnostics and healthcare.} Medical MMAD is shaped by strict requirements on localization granularity and clinical reliability: anomalies are subtle, high-stakes, and often demand pixel-/voxel-precise delineation (CH2--CH3). Multi-modal settings frequently combine 2D observations with 3D anatomy or longitudinal records, amplifying cross-modal alignment challenges (CH1) under severe data scarcity and subtype variability (CH4). Representative benchmarks span multi-modal ocular imaging, radiology, dermatology/endoscopy, and physiological monitoring~\cite{li2025adapting, porwal2020idrid, chilamkurthy2018deep, baid2021rsna, tbkk-q937-25, chowdhury2020can, codella2018skin, bernal2015wm, jha2019kvasir, hicks2021endotect, gong2021multi, moody2001impact, goldberger2000physiobank, johnson2016mimic}.

\mypara{Video and surveillance analytics.} VAD is commonly evaluated under weak supervision, where audio and language cues can reduce ambiguity in visually uncertain events (CH2) but introduce additional alignment and synchronization issues (CH1). Benchmarks span public-safety surveillance, audio-visual violence, multi-scenario settings, and event-based sensing, collectively stressing long-horizon modeling and granularity imbalance (CH3)~\cite{sultani2018real, luo2017revisit, wu2020not, ali2025avar, zhu2024advancing, qian2025ucf}.

\mypara{Cross-domain characteristics and benchmarking gaps.} \revise{Table~\ref{tab:all_datasets} suggests two recurring gaps. First, annotation \emph{granularity} differs sharply: industrial/medical tasks often need pixel-/voxel-level masks, while surveillance typically provides only frame-level labels. Second, most benchmarks remain overly \emph{stationary}; despite the importance of concept drift (CH5), few datasets reflect long-term non-stationarity as in WADI~\cite{ahmed2017wadi} or MIMIC-III~\cite{johnson2016mimic}.}

\revise{
\subsection{Evaluation Metrics and Protocols}
As MMAD spans diverse physical scenarios, a single unified evaluation metric is insufficient. While current evaluation \emph{metrics} are largely inherited from established unimodal domains to assess the final fused prediction, the emerging testing \emph{protocols} are increasingly tailored to probe unique cross-modal dynamics. The choice of metric is deeply coupled with the underlying modality and the specific \emph{granularity imbalance} (CH3) of the task. Furthermore, the rise of Foundation Models and generative paradigms has necessitated a shift from standard closed-set testing to more complex evaluation protocols.}

\revise{
\subsubsection{Muliti-granularity evaluation metrics.} To rigorously assess detection performance at different resolutions, domains use metrics aligned with their annotation granularity:}

\revise{
\begin{itemize}
\item \textbf{Spatial localization (image/3D):} In industrial inspection and medical imaging, evaluation should address both coarse detection and precise localization. \emph{Image-level AUROC} is the standard metric for sample-level decisions. For defect isolation, \emph{Pixel-level AUROC} and Area Under the Per-Region Overlap curve (\emph{AUPRO})~\cite{bergmann2019mvtec} are widely used; AUPRO is especially suitable for small, localized anomalies because it emphasizes region-wise overlap. In medical applications, overlap-based metrics such as the \emph{Dice score} and \emph{Intersection over Union (IoU)} are commonly reported to quantify structural consistency.   
\item \textbf{Spatio-temporal detection (video/audio):} For visual sequences and audio streams, anomalies stem from the \emph{temporal evolution} of patterns rather than a single frame. Although \emph{Frame-level AUROC} is widely used, mainstream benchmarks such as ShanghaiTech~\cite{liu2018future} increasingly report \emph{Average Precision (AP)}, which is more informative under severe class imbalance. \emph{Detection Latency} is also reported to measure how quickly a model responds after an anomalous event begins.
\item \textbf{Sequential dynamics (time-series/sensors):} In industrial sensing and IT monitoring, evaluation targets unimodal or multi-modal temporal dependencies, but metrics remain fragmented due to continuous temporal drift and ambiguous boundaries. Traditional \emph{Point-Wise (PW)} metrics underestimate performance when predictions are slightly time-shifted, while the widely used \emph{Point-Adjusted (PA)} F1-score~\cite{su2019robust} often overestimates it by rewarding partially detected segments. To address this, several alternatives have been proposed: \emph{Revised Point-Adjusted (RPA)}~\cite{hundman2018detecting} treats each anomalous segment as a single event, \emph{Affiliation metrics}~\cite{huet2022local} reward proximity to ground-truth intervals, and \emph{Volume Under the Surface (VUS)}~\cite{paparrizos2022volume} integrates accuracy over varying window sizes to reduce window-induced bias.
\end{itemize}
}

\revise{
\subsubsection{Emerging evaluation protocols.} 
Beyond static metrics, robust assessment of MMAD systems increasingly relies on protocol-level testing:
\begin{itemize}
\item \textbf{Zero-shot and open-set assessment:} With vision--language models, protocols increasingly evaluate \emph{zero-shot generalization}~\cite{jeong2023winclip, li2024promptad}, requiring models to detect unseen anomaly categories from text prompts and directly probing generalization under knowledge deficits (CH2).
\item \textbf{Fairness in anomaly injection:} Following recent work~\cite{li2021cutpaste, hu2024anomalydiffusion}, pseudo-anomaly synthesis protocols must prevent injected anomalies from leaking semantic labels or other shortcuts into training. Quantifying the synthetic-to-real distribution gap is thus a key fairness criterion.
\item \textbf{Modality-missing robustness:} Since real deployments rarely provide complete data streams, robustness protocols now measure performance degradation when a modality is missing at inference, testing the resilience of cross-modal alignment mechanisms~\cite{wang2023multimodal, wu2024vadclip} under severe modal discrepancy (CH1).
\end{itemize}
}

\vspace{-0.4em}

\revise{
\subsection{Computational Complexity and Scalability Analysis}
Foundation models and generative approaches have greatly advanced MMAD, but large-scale deployment is often limited by compute and memory. In Big Data settings—with high-resolution inputs, long time spans, and continuous multi-modal streams—scalability becomes a primary concern, not a secondary one after accuracy.}

\revise{
From a system perspective, two dominant sources of computational overhead are commonly observed:
\begin{itemize}
\item \textbf{FM-based cross-modal alignment:} Many recent MMAD methods use high-capacity Transformers to align heterogeneous modalities~\cite{radford2021learning,li2024promptad, cao2024adaclip}. Standard self-attention can incur up to $\mathcal{O}(N^2)$ computation and memory in sequence length $N$, so jointly processing high-resolution visual tokens with synchronized text or 3D representations can cause high GPU memory use and latency, limiting throughput in large-scale deployments.
\item \textbf{Generative and diffusion-based paradigms:} Generative methods, especially diffusion-based anomaly synthesis~\cite{hu2024anomalydiffusion}, require iterative denoising over $T$ steps, yielding overall complexity approximately proportional to
$\mathcal{O}(T \cdot C_{\text{step}})$,
where $T$ denotes the denoising steps and
$C_{\text{step}}$ is the cost of one network evaluation. Although they produce diverse pseudo-anomalies, these multi-step procedures can incur substantial overhead, particularly for online generation or adaptation.
\end{itemize}
}
\revise{
Beyond these backbone costs, cross-modal interaction adds burden because aligning heterogeneous representations often needs repeated fusion or joint encoding, with overhead especially high in streaming and high-throughput settings. Computational limits are not just system concerns but can directly reduce detection quality: efficiency demands often force lower spatial resolution or temporal sampling, which, though scalable, can reduce sensitivity to fine-grained or brief anomalies and worsen the granularity imbalance discussed in CH3.}

\revise{
To better support real-world deployment, scalable MMAD systems should explicitly navigate the accuracy--efficiency trade-off frontier. Potential directions include \emph{model compression} (e.g., knowledge distillation~\cite{gu2024rethinking}) and \emph{parameter-efficient adaptation}~\cite{li2024promptad,cao2024adaclip}, which aim to reduce overhead while preserving cross-modal alignment capability.
}

\section{Open Problems and Future Directions}
\label{sec:future}

MMAD is increasingly shaped by multi-modal foundation models and hybrid pipelines that couple normality modeling with anomaly-oriented perturbations. Yet core challenges persist, including cross-modal misalignment (CH1), ambiguous boundaries (CH2), imbalanced sensitivity (CH3), a synthetic-to-real gap (CH4), and limited adaptability under drift (CH5).
\revise{We condense future directions into one unifying paradigm and three core research questions (RQs).}

\revise{\textbf{Unifying paradigm: Foundation-model-driven assumption fusion.}
Rather than separating normality modeling and abnormality injection, future MMAD frameworks should \emph{leverage} foundation models as building blocks to couple both: use their priors to align normal representations, and generate targeted semantic edits/prompts as hard negatives to tighten the boundary. 
In parallel, rigorous theory for such pipelines (e.g., information-theoretic bounds for fusion and generalization under drift) remains limited and is an important open frontier.}

\revise{\textbf{RQ1: 
How to generate reliable anomaly hypotheses?
}
Pseudo anomalies can tighten boundaries, but their realism is rarely measurable; with scarce real anomalies, metrics such as FID may overfit. Reference-free protocols (auxiliary discriminators, alignment checks, and physics/temporal constraints) are needed to validate pseudo anomalies across modalities.}

\revise{\textbf{RQ2: 
How to perform semantic reasoning under cross-modal conflicts?}
To mitigate modality dominance and conflicting judgments, MMAD needs uncertainty-aware fusion such as evidential fusion or dynamic blending, plus semantic grounding. Foundation models, particularly MLLMs, can map anomaly evidence to human-readable rationales by linking continuous representations to violated cross-modal constraints.}

\revise{\textbf{RQ3:
How to continually adapt under evolving environments?
}
Real-world MMAD systems must continually adapt to evolving distributions while maintaining stable decision boundaries under constrained resources.
Promising directions include PEFT adapters with replay/prototype retention, linear-time sequence models as alternatives to $\mathcal{O}(N^2)$ attention, and token pruning/distillation for real-time operation.}

\section{Conclusions}
\label{sec:conclusion}
This survey provided a structured, assumption-driven review of multi-modal anomaly detection, organizing prior work around two complementary paradigms: normality modeling and anomaly injection. By focusing on underlying assumptions rather than specific architectures or modalities, we offered a unified lens that clarifies both the strengths and limitations of existing approaches.
We highlighted that MMAD complexity stems from the interplay between intrinsic anomaly detection challenges and multimodality-specific difficulties, particularly heterogeneity. Within this framework, we examined how current methods tackle these issues across multiple levels, from representation learning and structural perturbations to semantic injection and knowledge-guided modeling.
Overall, MMAD is shifting from isolated detection pipelines toward holistic perception systems that combine representation learning, anomaly simulation, semantic reasoning, and deployment-aware adaptation. Foundation models are emerging as a key catalyst, reducing modal heterogeneity and enabling knowledge-rich anomaly definitions. By synthesizing recent progress and articulating actionable open questions, we aim for this survey to serve as both a conceptual reference and a roadmap toward adaptive, interpretable, and reliable real-world MMAD systems.

\bibliographystyle{IEEEtran}
\bibliography{ref.bib}

\newpage

\begin{IEEEbiography}
[{\includegraphics[width=1.1in,height=1.25in,clip,keepaspectratio]{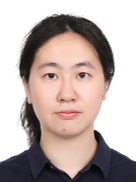}}]{Xudong Mou}
received her M.S. degree in the School of Computer Science and Engineering, Beihang University, in 2021. She is working towards a Ph.D. at the School of Computer Science and Engineering, Beihang University, China. Her research interest is time series anomaly detection.\end{IEEEbiography}
\vskip -4\baselineskip

\begin{IEEEbiography}[{\includegraphics[width=1in,height=1.25in,clip,keepaspectratio]{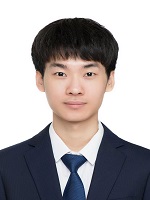}}] {Zexin Wu} received his B.S. degree in Electrical Engineering and Automation from China University of Geosciences, Beijing, in 2020. He is working towards an M.S. at the School of Software, Beihang University, China. His research interest is time series anomaly detection.
\end{IEEEbiography}
\vskip -4\baselineskip

\begin{IEEEbiography}[{\includegraphics[width=1.1in,height=1.25in,clip,keepaspectratio]{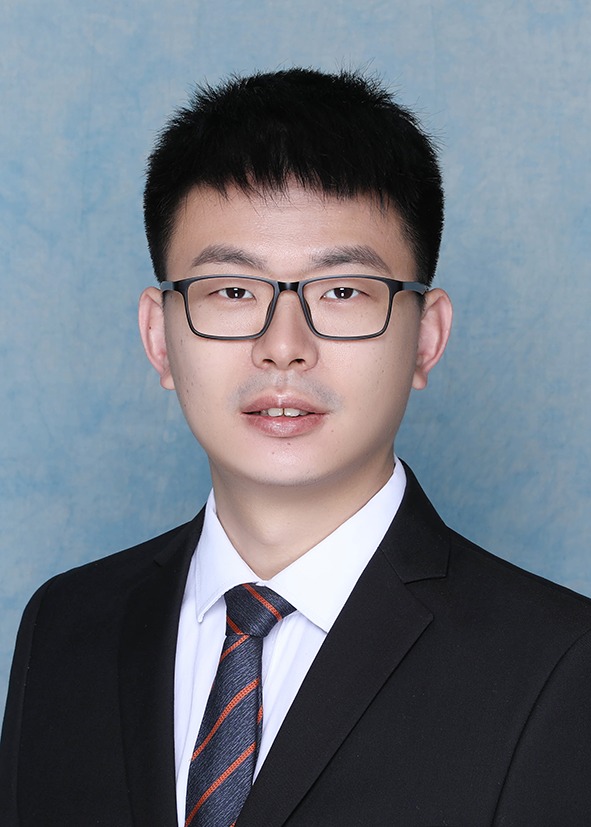}}]{Chuan Luo} received the Ph.D. degree in computer science from Peking University, Beijing, China, in 2016. He is currently an associate professor with School of Software, Beihang University, Beijing, China. His current research interests include constraint solving and heuristic search.\end{IEEEbiography}

\vskip -4\baselineskip
\begin{IEEEbiography}[{\includegraphics[width=1.1in,height=1.25in,clip,keepaspectratio]{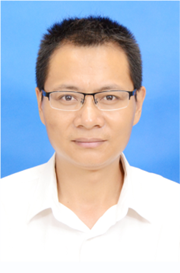}}]{Shiru Chen}
    received a PhD degree from Harbin Engineering University. He is currently a R\&D Director of Shandong Inspur Intelligent Production Technology Co., Ltd. and has over 20 years of experience in the fields of communication, IoT, AI, distributed data processing, and industrial software systems. He has published 25+ academic papers, holds 60 invention patents, participated in 5 national standards writing, and authored 2 books. \end{IEEEbiography}

\vskip -4\baselineskip
\begin{IEEEbiography}[{\includegraphics[width=1.1in,height=1.25in,clip,keepaspectratio]{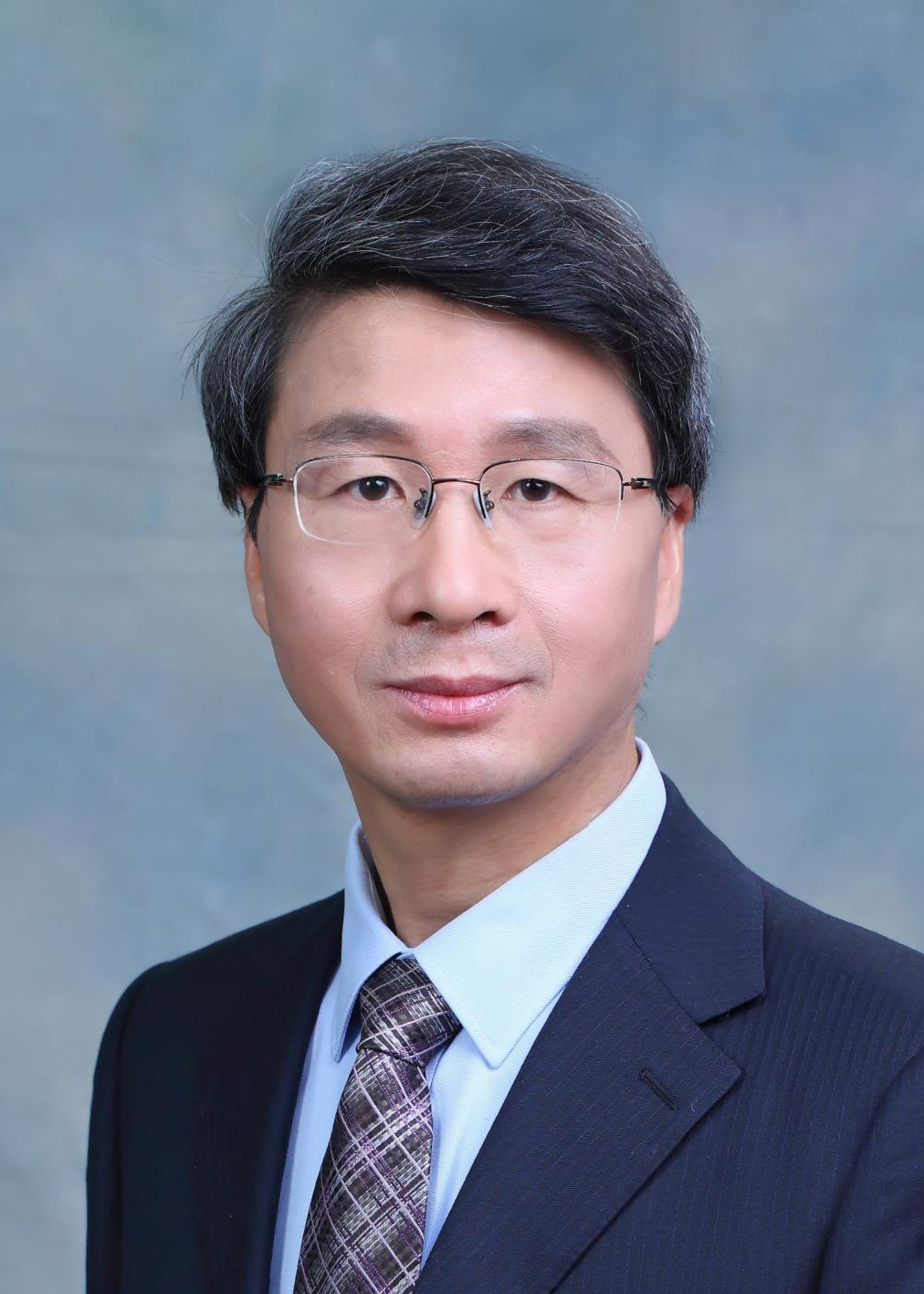}}]{Xudong Liu}
is a professor at the School of Computer Science and Engineering, Beihang University, Beijing, China. He has led several China 863 key projects and government projects. His research interests include software middleware technology, software development methods and tools, large-scale information technology projects, and the application of research and teaching.\end{IEEEbiography}
\vskip -4\baselineskip

\begin{IEEEbiography}[{\includegraphics[width=1.1in,height=1.25in,clip,keepaspectratio]{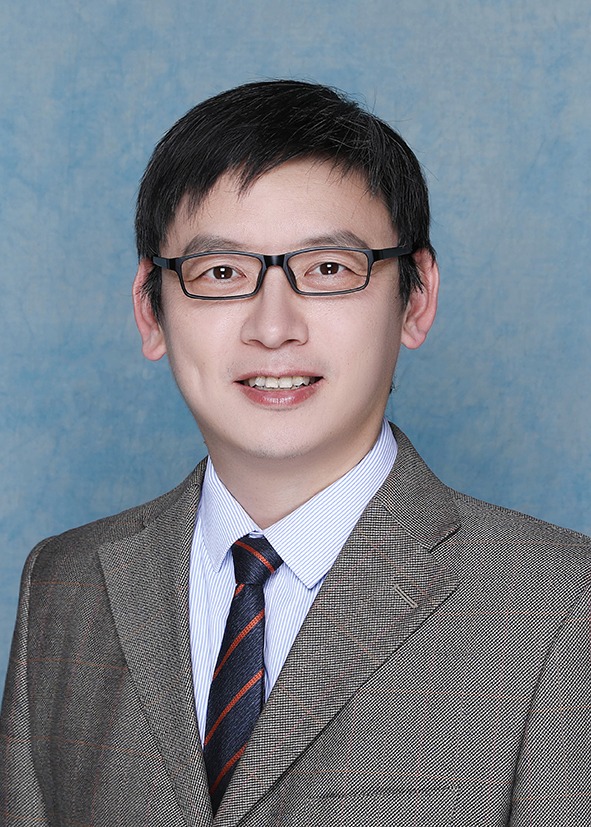}}]{Chunming Hu} received the Ph.D. degree in computer science from Beihang University, Beijing, China, in 2006. Currently, he is a Professor with the School of Software, Beihang University, Beijing, China. His research interests include distributed systems, system virtualization, data management, and processing systems.
\end{IEEEbiography}
\vskip -4\baselineskip

\begin{IEEEbiography}[{\includegraphics[width=1in,height=1.25in,clip,keepaspectratio]{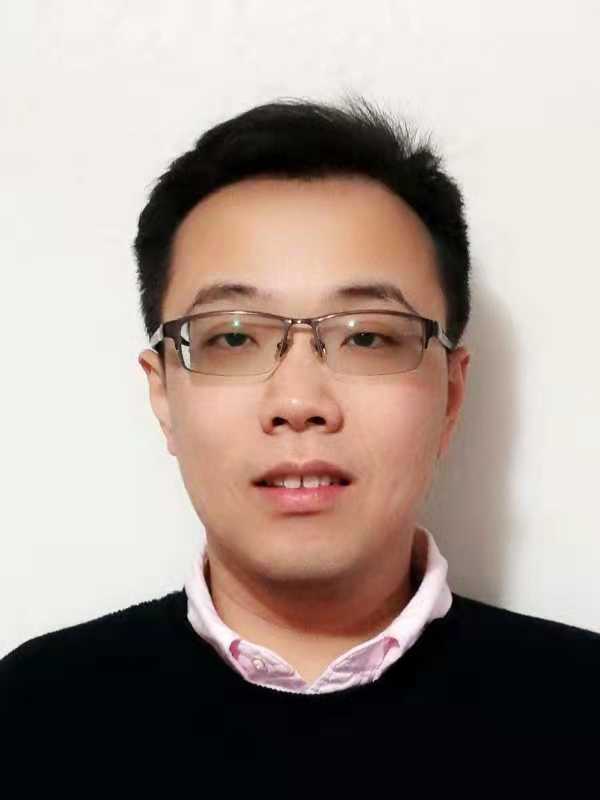}}] {Renyu Yang} is an associate professor with the School of Software, Beihang University, China. Prior to this, he was with the University of Leeds UK, Alibaba Group China and Edgetic Ltd. UK, building large-scale computing/AI infrastructures. He is a recipient of Alan Turing Post-Doctoral Enrichment Award, 2022. His research interests include parallel and distributed computing, large-scale AI systems and software dependability. He is a member of IEEE.
\end{IEEEbiography}

\end{document}